\documentclass[10pt,letterpaper]{article}
\usepackage[top=0.85in,left=2.75in,footskip=0.75in]{geometry}

\usepackage{amsmath,amssymb}

\usepackage{changepage}

\usepackage{textcomp,marvosym}

\usepackage{cite}

\usepackage{nameref,hyperref}

\usepackage[right]{lineno}

\usepackage[nopatch=eqnum]{microtype}
\DisableLigatures[f]{encoding = *, family = * }

\usepackage[table]{xcolor}

\usepackage{array}
\usepackage{booktabs} 

\newcolumntype{+}{!{\vrule width 2pt}}

\newlength\savedwidth

\raggedright
\usepackage[aboveskip=1pt,labelfont=bf,labelsep=period,justification=raggedright,singlelinecheck=off]{caption}

\makeatletter
\renewcommand{\@biblabel}[1]{\quad#1.}
\makeatother

\usepackage{lastpage,fancyhdr,graphicx}
\usepackage{epstopdf}
\fancyheadoffset[L]{2.25in}
\fancyfootoffset[L]{2.25in}
\begin{document}
\vspace*{0.2in}

\begin{flushleft}
{\Large
\textbf{Attention to task structure for cognitive flexibility} 
}
\newline
\\
Xiaoyu K. Zhang\textsuperscript{1*},
Mehdi Senoussi\textsuperscript{2}
Tom Verguts\textsuperscript{1},
\\
\bigskip
\textbf{1} Department of Experimental Psychology, Ghent University, Henri Dunantlaan 2, 9000 Gent, Belgium
\\
\textbf{2} Univ Toulouse, Université Toulouse Jean Jaurès, CNRS, CLLE, Toulouse, France
\\

\bigskip

* xiazhan.zhang@ugent.be

\end{flushleft}


%
\section*{Abstract}
Humans and artificial agents must often learn and switch between multiple tasks in dynamic environments. Success in such settings requires cognitive flexibility: the ability to retain prior knowledge (cognitive stability) while also transferring it to novel tasks (cognitive generalization). Cognitive flexibility research has largely focused on the role of model architecture to achieve these complementary goals. However, it is less well understood how the structure of the environment itself influences cognitive flexibility, and how it interacts with model architecture.
To address this gap, we design a multi-task learning environment in which tasks are defined by a combination of two cue dimensions, allowing us to characterize the environment with graph-theory methods. We also introduce gating-based (multiplicative) and concatenation-based attention models that can decompose tasks into components and can sequentially allocate attention to them. We compare the attention-based models’ performance in the multi-task learning environment to multilayer perceptrons. Generalization and stability are systematically evaluated across environments that vary in richness and task connectivity.
We observe that richer environments improve both generalization and stability. In addition, a critical novel observation is that (graph theory based) connectivity between the tasks in the environment strongly modulates both stability and generalization, with especially pronounced benefits for attention-based models. These findings underscore the importance of considering not only cognitive architectures but also environmental structure and their interaction in shaping multi-task learning, generalization, and stability.


%
\section*{Author summary}
Throughout our lives, humans are confronted with one task after another. Fortunately, we don't have to learn each one anew. Almost all tasks allow reusing components that were useful in earlier ones. For example, when learning to ride a scooter, one can reuse traffic rules that were learned while riding a bike. Efficient learning requires learning to reuse those earlier components. In this way, one can both learn novel tasks quickly while at the same time retaining the knowledge to perform earlier tasks. Here, we develop a class of attention-based models and compare them with standard multi-layer perceptrons in solving a class of sequentially, componentially related tasks. We find that the attention-based models are better in extracting componential structure, and particularly in well-structured environments. This result means that not just the cognitive architecture matters for learning, but also how it fits to its environment. Stated otherwise, attention is not all you need: it also needs to fit its environment.   

\section*{Introduction}

Biological and artificial agents operate in dynamic environments where they must learn, perform, and switch between multiple tasks. This creates both opportunities and challenges for how knowledge is extracted from the environment, stored, and shared across tasks.

A central opportunity in multi-task learning is the potential to identify and reuse common components across tasks, enabling generalization. At the same time, multi-task learning introduces a major challenge: interference between tasks, impairing stability. We refer to the combination of generalization and stability as cognitive flexibility: the ability to transfer knowledge to novel tasks while maintaining performance on previously learned tasks.

Notoriously, neural networks (the computational tool to study both biological and artificial agents), are highly susceptible to catastrophic forgetting. In particular, previously acquired knowledge tends to be overwritten when new tasks are learned~\cite{de_lange_continual_2023, french_catastrophic_1999, grossberg_how_1980,9349197,MCCLOSKEY1989109,kim2023stabilityplasticitydilemmaclassincrementallearning,mcclelland1995there}. This fact relates to a fundamental generalization-stability tradeoff: models that protect against interference do so via conjunctive representations, which achieve stability, but at the cost of sacrificing generalization, as conjunctive representations do not allow generalization across overlapping tasks ~\cite{lippl2025when}. Instead, models that generalize typically have compositional (instead of conjunctive) representations, but at the risk of interference between tasks~\cite{musslick2021rationalizing}. A natural route to generalization, therefore, is to decompose tasks into reusable components so that knowledge learned in one context can be recombined in another~\cite{correa_humans_2023}. However, due to the generalization-stability tradeoff, decomposition introduces a challenge: when multiple tasks share partially overlapping components, the agent must decide which components to reuse and which to keep separate; otherwise, shared representations can induce interference across tasks. 

Several approaches have been proposed to mitigate interference and improve stability in multi-task learning. A first approach is regularization-based continual learning, which penalizes parameter updates that would increase the loss on earlier tasks~\cite{kirkpatrick2017overcoming}. A second approach is replay or interleaved training, which periodically retrains a model with earlier task information to preserve performance ~\cite{mcclelland1995there,DBLP:journals/corr/ShinLKK17}. A third major class of methods relies on weight protection, in which parameters supporting earlier tasks are selectively constrained to prevent overwriting ~\cite{grossberg_how_1980,verbeke_learning_2019}. Conjunctive representations can be viewed as one instance of this strategy: by encoding task-specific combinations, they promote stability, but can also limit generalization across overlapping tasks. More recently, researchers have explored attention and gating mechanisms as a dynamic form of protection. Here, task-relevant pathways are selectively amplified to reduce cross-task interference. This improves continual learning without relying solely on static, global weight constraints~\cite{hummos23,sommers25,VERBEKE2022256}.

However, beyond architecture, the structure of the environment is equally critical. Recent works show that environmental properties shape the representations that emerge. For instance, modularity arises in the hidden representations when the task environment is sufficiently varied~\cite{dorrell2025rangeindependencedrivesmodularity,johnston2023abstract}. Also the training curriculum matters: how training examples are ordered can substantially determine whether agents acquire reusable structure and thus generalize ~\cite{dekker_curriculum_2022}. Task structure can also influence catastrophic forgetting: Both very dissimilar and very similar tasks are not prone to interference, whereas partially overlapping tasks are~\cite{holton2026humans,saxe19,lee2022maslow}. Yet, the global task connectivity structure of the environment remains comparatively underexplored.

In this work, we investigate how generalization can combine with stability in modularly structured environments by models that exploit such structure. We introduce a general multi-task space where each task is defined as a combination of sensory and motor dimensions. From this space, we sample a subset of tasks to form a regime, and systematically manipulate two key properties of the regime. The first is richness, which refers to the number of components available for constructing tasks. The second is connectivity, which refers to the degree to which tasks share components, yielding a structured component-connected graph. Models are first trained on tasks from the regime and evaluated on novel tasks that recombine familiar components in novel ways, providing a measure of generalization. We then train models on these novel tasks and assess performance on the original regime tasks, quantifying stability in the face of potential interference.

To examine how architecture shapes learning under these environmental constraints, we compare standard multilayer perceptrons (MLPs) with a novel attention-augmented multilayer architecture designed to simulate human-like decomposition. This architecture introduces multiple attention layers that encourage task decomposition by selectively routing task-relevant information and focusing on different components in a sequential manner. Attention is implemented either through gating, which modulates the stimulus processing information flow, or through concatenation, which integrates selected features into the stimulus processing stream into a joint representation (see Methods). The attention selection process is learned via gradient descent jointly with the stimulus processing itself, enabling the network to sequentially focus on different components and support flexible recombination across tasks. We further examine how environmental richness and connectivity shape the emergence of cue-sensitive representations and whether differences between MLPs and attention-based models can be explained by cue sensitivity.

Our results show that richer and more connected task environments improve both generalization to novel tasks and stability on previously learned tasks, with particularly strong gains for attention-based models. In rich environments, attention-based models develop layer-wise cue-sensitive representations, suggesting a potential mechanism by which selective routing can mitigate the generalization–stability tradeoff. Together, these findings highlight that multi-task learning is shaped not only by model architecture, but also by the richness and connectivity structure of the task environment and its interaction with the learner.

\section*{Results}
\subsection*{Multi-task structure}
\begin{figure}[!h]
\includegraphics[width=\textwidth]{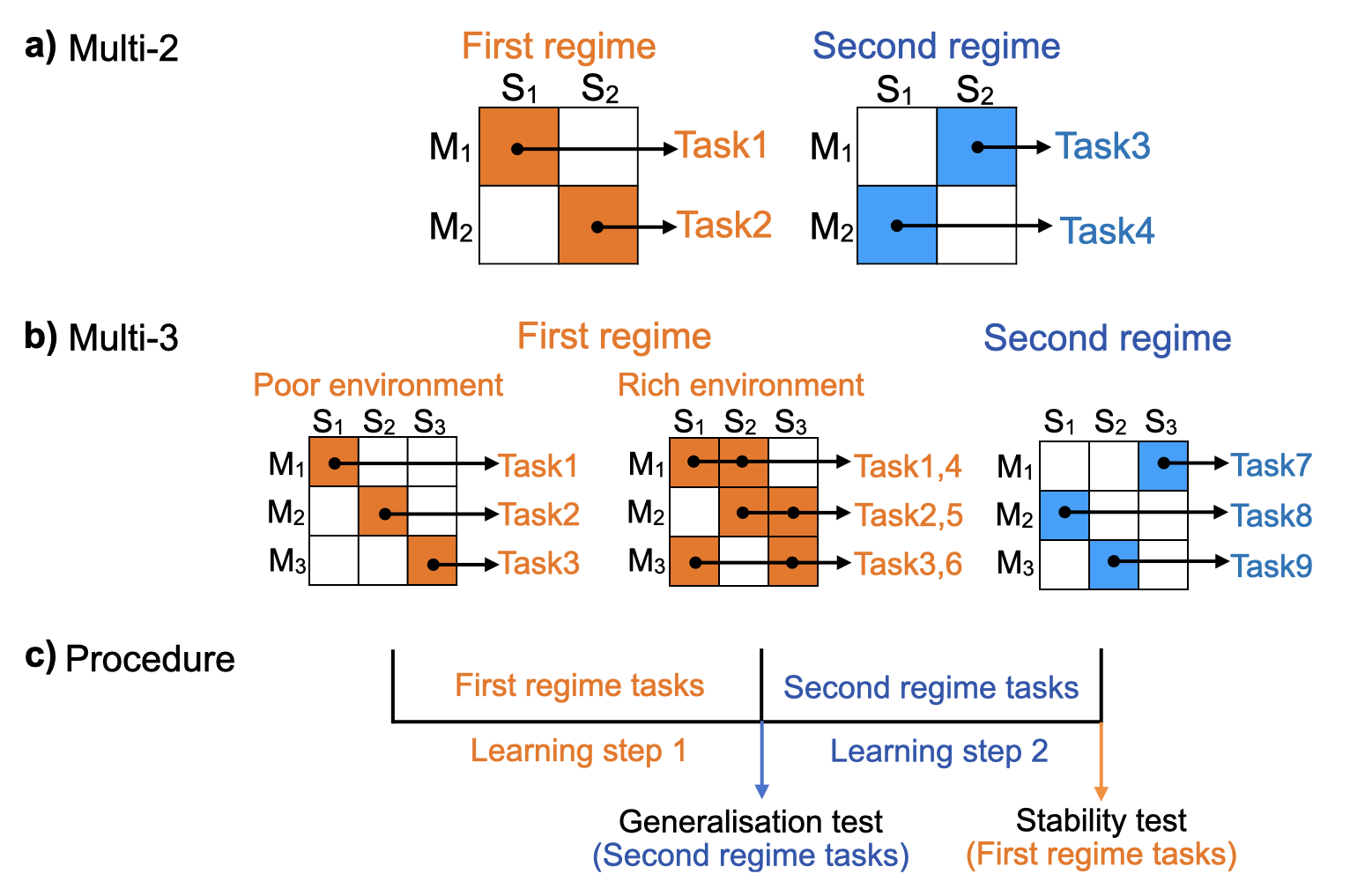}
\caption{\textbf{Overview of the multidimensional task structure.}
(a) Multi-2: A $2 \times 2$ dimensional task structure with two sensory cues $(S_1, S_2)$ and two motor cues $(M_1, M_2)$. There are four combinations of cues (4 tasks) in the task structure. The first regime includes two tasks, $(S_1, M_1)$ and $(S_2, M_2)$, while the second regime contains two novel tasks, $(S_1, M_2)$ and $(S_2, M_1)$. (b) Multi-3: A $3 \times 3$ dimensional task structure with three sensory cues and three motor cues. The first regime includes three tasks in a poor environment and six tasks in a rich environment. The second regime is identical for both, featuring three novel tasks. (c) Procedure: First regime tasks are learned with feedback, followed by a generalization test on second regime tasks without feedback. After second regime tasks with feedback, a stability test is conducted using the original first regime tasks without feedback.}
\label{fig1}
\end{figure}

To simulate the multi-task environment, a multi-dimensional structure, called Multi-$n$, is built with two modalities (sensory and motor), each modality consisting of $n$ dimensions. The sensory modality has dimensions $S_1, \ldots, S_i, \ldots, S_n$ such as color, shape, or size. Each dimension has two possible values. For example, the color dimension may take values red and green. The motor modality has dimensions (or effectors) $M_1, \ldots, M_i, \ldots, M_n$ such as index, middle, or ring finger. Here also, each dimension has two possible values: For example, values on the index finger dimension can be left or right (i.e., one can press the left or right index finger). The dimensionality of the task structure is determined by the number of sensory and motor cues, each corresponding to a sensory or motor dimension. For example the task structure with 2 sensory cues and 2 motor cues is called Multi-2 (see Fig~\ref{fig1}a). Each combination of one sensory and one motor cue represents a task, for example task 1 (color, index finger) and task 2 (shape, middle finger) (see Fig~\ref{fig1}a). On each trial, a sensory cue specifies the relevant sensory dimension of the stimulus (e.g., color or shape), while a motor cue designates the motor effector to respond (e.g., index or middle finger).

In general, in a Multi-$n$ structure where each sensory dimension can take on two possible values (e.g., red or blue for the color dimension), there are $2^n$ possible stimuli. The task determines the mapping of each stimulus to a specific response. The abstract mapping is the same across motor dimensions. For instance, if the sensory cue is color and the stimulus is red, the left response could be the correct one. The motor cue then indicates which motor effector must be used to provide that response (e.g., index finger). After each response, feedback is provided during learning, guiding the model to learn the initially unknown task-relevant mappings between cues, stimuli and responses.

\subsection*{Model structure}

The models we compare in this study include standard multilayer perceptrons (MLPs) and attention-based architectures (Fig~\ref{fig14}). The MLP baseline consists of sequentially stacked fully connected (dense) layers (see Fig~\ref{fig14}a). In contrast, the attention-based models augment an MLP backbone with one of two types of attention layers: Attention-Gating or Attention-Concatenation. In the Attention-Gating variant, attention layers apply multiplicative gates to the outputs of MLP layers, selectively filtering stimulus features based on task-relevant information. In the Attention-Concatenation variant, attention layers concatenate intermediate (MLP) representations, allowing the model to integrate multiple feature streams and potentially capture relations across representations more explicitly.

Like the MLP baselines, attention-based models use multiple dense layers to process the stimulus input. These layers are labeled Dense1, Dense2, etc. (see Methods, Fig~\ref{fig14}, for details). However, unlike MLPs, they include a dedicated cue-processing stream that encodes task cues. The resulting cue representation is used to guide information selection in the stimulus-processing stream through the attention layers. Importantly, Dense1 encodes the task cues through a dense transformation, allowing the model to autonomously learn which cue features should modulate attention at each attention layer. We include two attention layers, motivated by the hypothesis that the model may learn to select sensory-relevant and motor-relevant components of the task structure at different stages of processing.

Beyond architectural differences, each model type includes two variants. For MLPs, we vary depth: We build MLP\_1 and MLP\_2, where MLP\_2 has one additional hidden layer compared to MLP\_1. For attention-based models, we match the backbone depth of MLP\_1 and manipulate whether information between the two attention layers passes through a bottleneck (a lower-dimensional hidden layer), which constrains representational capacity and may encourage more efficient or factorized intermediate representations. Specifically, Gate\_1 and Concat\_1 include a bottleneck, whereas Gate\_2 and Concat\_2 don’t.

\subsection*{Multi-2 \& Multi-3 environments}

The Multi-2 environment has 4 different tasks with 4 combinations of sensory and motor cues (Fig~\ref{fig1}a). We separate those 4 tasks into 2 non-overlapping sets of tasks to be trained consecutively, called the first (training) regime and second regime. In Multi-2 and Multi-3, each (sensory and motor) cue appears at least once in each regime. The models are initially trained with the first regime with feedback (learning step 1; Fig~\ref{fig1}c). The second regime is used to test if the models can transfer the information learned in the first regime and do the novel tasks without learning (without feedback), which we call a generalization test. After that, the models are trained on the second regime with feedback (learning step 2) and then tested with the first regime without feedback to see if the models can remember the tasks they learned before, which we call a stability test. Multi-3 is constructed similarly (Fig~\ref{fig1}b).

To explore how environmental structure affects generalization and stability, we investigate the effect of richness (see Fig~\ref{fig1}b). Because of the limitation on the number of tasks in Multi-2, we do this only in Multi-3, in which the first regime can have 3 tasks in the poor environment and 6 tasks in the rich environment (Fig~\ref{fig1}b). At each richness level, the first regime always includes all cue information, that is, all values $S_i$ (e.g., color, shape, orientation), and all values $M_i$ (e.g., index, middle, and ring finger). Otherwise, the absence of one cue (for example, color) would significantly impair the model’s generalization ability, as it would fail to generalize to any task involving that cue.

\subsubsection*{Learning curves}

\begin{figure}[!h]
\includegraphics[width=\textwidth]{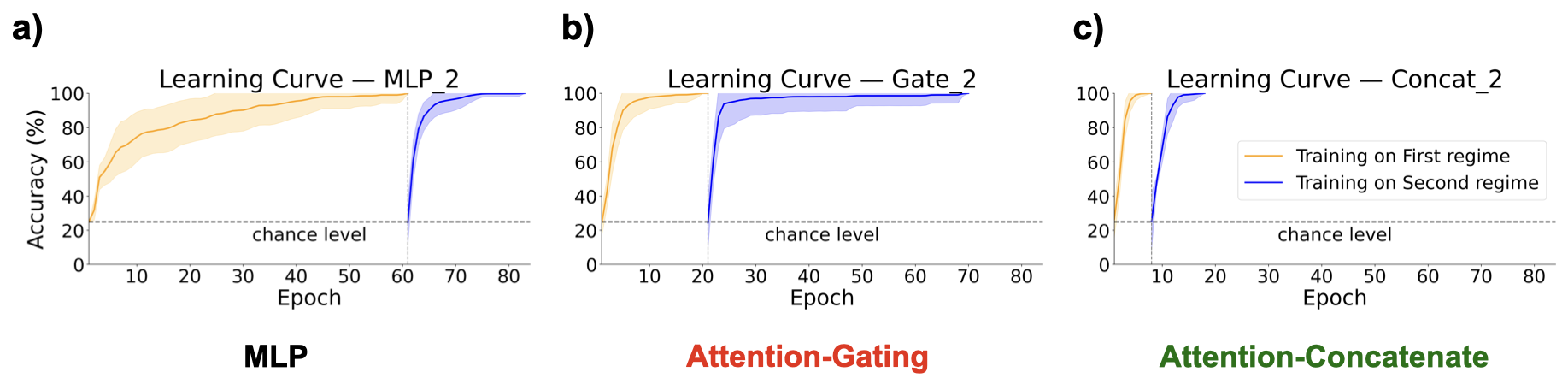}
\caption{\textbf{Learning curves in Multi-2 for three model types.}
Curves show mean training accuracy over 50 runs for (a) MLP\_2, (b) Gate\_2, and (c) Concat\_2, representing the MLP, Attention-Gating, and Attention-Concatenate architectures, respectively. Orange traces correspond to learning step 1 (training on the first-regime tasks), and blue traces correspond to learning step 2 (training on the second-regime tasks). Shaded regions indicate ±1 standard deviation across runs (cropped at 100\%). The vertical dashed line marks the transition between learning regimes.}
\label{fig2}
\end{figure}

All models were trained and tested in three environments: Multi-2, Multi-3 Poor, and Multi-3 Rich. In each environment, tasks were organized into regimes, with the number of tasks per regime varying by environment. For every task, we generated 5{,}000 trials; trials from all tasks within a regime were randomly shuffled to form the training set. Accordingly, one epoch corresponds to one full pass through the regime’s shuffled dataset (i.e., $5{,}000 \times$ number of tasks in the regime trials). Training terminated via early stopping once the model reached 100\% accuracy for four consecutive epochs. To evaluate the variability in model performance, we implemented 70 independent runs, of which we chose the first 50 that reached $>99\%$ accuracy. We report mean accuracy across runs. Learning curves for training step 1 and 2 are shown in Fig~\ref{fig2}, and summary measures of generalization and stability are reported in Fig~\ref{fig3}a-b, respectively. 

Fig~\ref{fig2} plots learning curves in the Multi-2 environment for three model types: an MLP, an Attention-Gating model, and an Attention-Concatenate model. Because learning trajectories were nearly identical within each model type, we show one representative model per type (MLP\_2, Gate\_2, and Concat\_2). For clarity, all plots here and below will show results for those three models; full results for all models are provided in the Appendix (except Fig~\ref{fig7}~\ref{fig10}~\ref{fig11} which show results for all models). All three model types achieved near-ceiling accuracy on the first-regime tasks by the end of step 1 ($\approx 100\%$) and on the second-regime tasks by the end of step 2 ($\approx 100\%$), indicating successful learning of all tasks in Multi-2 under the two-step training procedure.

\subsubsection*{Generalization and stability in environments with different richness
}

\begin{figure}[!h]
\includegraphics[width=\textwidth]{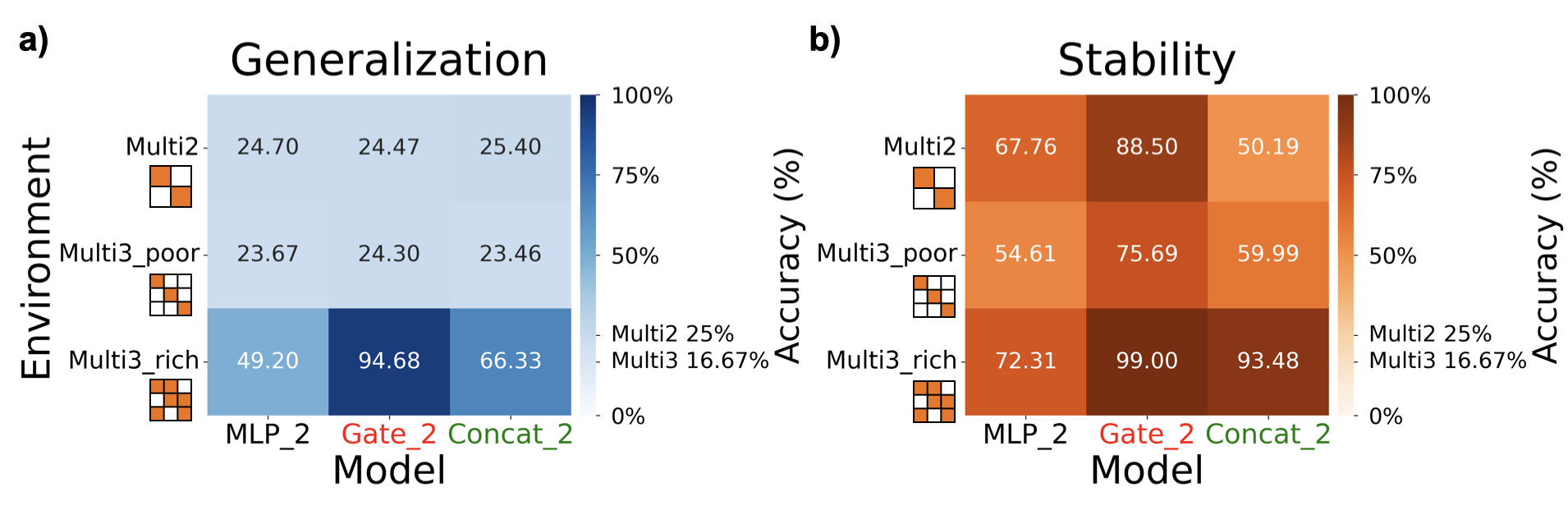}
\caption{\textbf{Performance of models in generalization and stability across different environments (Multi-2, Multi-3 Poor, Multi-3 Rich).}
Mean accuracy (averaged over 50 runs) for MLP\_2, Gate\_2, and Concat\_2. (a) Generalization results show higher accuracy for attention-based models compared to the MLP model, with near-perfect performance in the rich environment. (b) Stability results highlight the robustness of attention-based models in retaining prior knowledge, while the MLP model exhibits catastrophic forgetting. All models perform better in the rich than poor environment in both generalization and stability. The chance level for each environment is shown on the color bar.}
\label{fig3}
\end{figure}

Fig~\ref{fig3} shows generalization and stability in the Multi-2, Multi-3 Poor, and Multi-3 Rich environments for three model types: an MLP, an Attention-Gating model, and an Attention-Concatenate model. Here and below, we use the convention to indicate the minimum–maximum spread in accuracy across runs (i.e., the accuracy range). Considering generalization (Fig~\ref{fig3}a), in the Multi-2 and Multi-3 Poor, the performance of the MLP model (23.67--24.70\% accuracy range) and attention-based models (23.46--25.40\%) is comparable. However, in the Multi-3 Rich environment, attention-based models perform much better than the MLP model (MLP\_2: 49.20\%; attention-based models: 66.33--94.68\%). 
 
A similar conclusion holds for stability (see Fig~\ref{fig3}b). The MLP model exhibits pronounced catastrophic forgetting, most notably in Multi-2, where its stability (67.76\%) is below that of the Attention-Gating model (88.50\%) and comparable to the Attention-Concatenate model (50.19\%). In Multi-3 Poor, MLPs show low retention (54.61\%), while all attention-based models retain more knowledge (59.99--75.69\%), indicating reduced forgetting. In Multi-3 Rich, MLP model stability improves compared to the poor environment. Attention-based models remain markedly stronger: Attention-Gating models reach near-ceiling stability, and Attention-Concatenate models maintain high performance, indicating better preservation of previously acquired knowledge. 

Overall, these results show that increasing environmental richness improves performance for all models. However, attention-based models consistently generalize better than MLPs in both poor and rich environments, suggesting that attention mechanisms—especially gating—provide a clear robustness advantage by delivering stronger retention and greater resistance to forgetting than the MLPs.

\subsection*{Multi-4 environments}

\begin{figure}[!h]
\includegraphics[width=\textwidth]{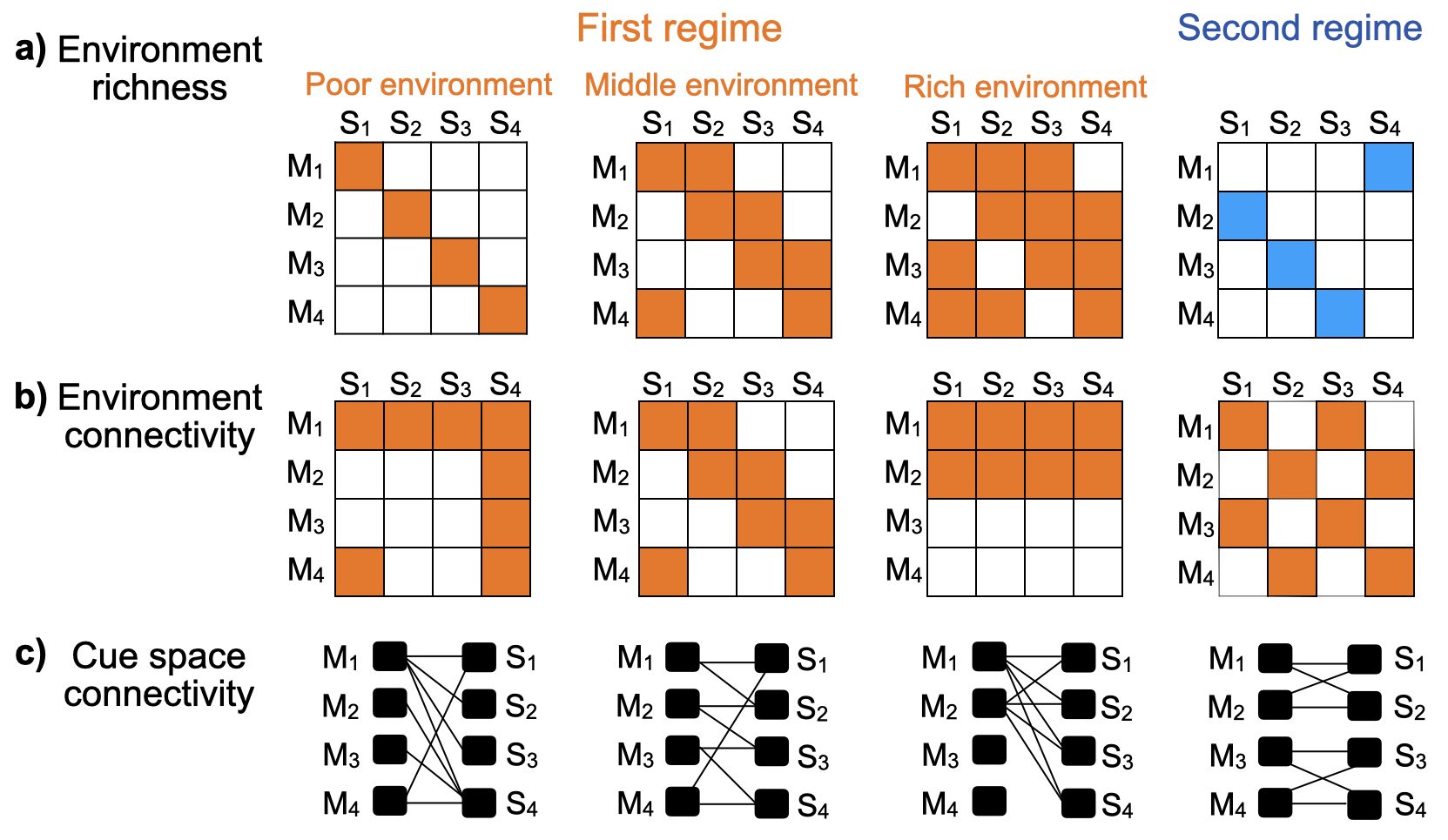}
\caption{\textbf{Experimental design for Multi-4 and connectivity graphs.}
Mean accuracy (averaged over 50 runs) for MLP\_2, Gate\_2, and Concat\_2. (a) Environment richness. In Multi-4, the first regime varies by environmental richness: poor (4 orange tasks), middle (8 orange tasks), and rich (12 orange tasks). The second regime tasks (4 blue tasks) consist of tasks that are different from all first regime tasks. (b) Environment connectivity. Different first regimes are shown for the middle environment (8 orange tasks), that vary in their connectivity. The second regime tasks for each first regime consists of the remaining eight tasks (shown in white) that are not included in the first regime. (c) Cue space connectivity. Connectivity of first regimes from (b) in cue space. Each cue, $M$ or $S$, is represented as a vertex and each edge between a pair of vertices represents a task. The two diagrams on the left illustrate connected regimes (all vertices are connected through a sequence of edges), while the two diagrams on the right illustrate disconnected regimes (not all vertices are connected).}
\label{fig4}
\end{figure}

Multi-4 (Fig~\ref{fig4}a) enables a more complete analysis of environmental structure by extending the Multi-3 design and adding a middle condition. We define three richness levels for the first regime: poor (4/16 tasks), middle (8/16 tasks), and rich (12/16 tasks). The middle condition supports multiple possible task combinations in the first regime, allowing a more detailed analysis of task-structure effects. Across all conditions, the second regime consists of the tasks not included in the first regime—shown as blue tasks in Fig~\ref{fig4}a. For the middle first-regime examples in Fig~\ref{fig4}b, the corresponding second regimes are the white tasks in each panel (and thus differ across regimes). To investigate this structure, we represented the first regime tasks via graphs (see Fig~\ref{fig4}c and Methods). In this way, the first regimes can be classified as either connected or disconnected regimes (loosely, can one “travel” from any cue to any other cue in the regime;  e.g., in Fig~\ref{fig4}b, regimes 1 and 2 are connected; regimes 3 and 4 are disconnected). Besides connectedness (binary variable, connected or not), we calculate the strength of connectivity (continuous variable) among tasks in the connected regimes, either via the average shortest path length (ASPL) or the longest shortest path length (LSPL) metrics from graph theory~\cite{watts1998collective}. ASPL is the average of the shortest distance between all pairs of cues. It quantifies how efficiently information can flow between tasks. LSPL, or graph diameter, captures the maximum distance between any two cues, indicating the overall extent or “spread” of the regime. Together, these metrics characterize how tightly or loosely integrated the connected task regime is (see Methods).

\subsubsection*{Learning curves}

\begin{figure}[!h]
\includegraphics[width=\textwidth]{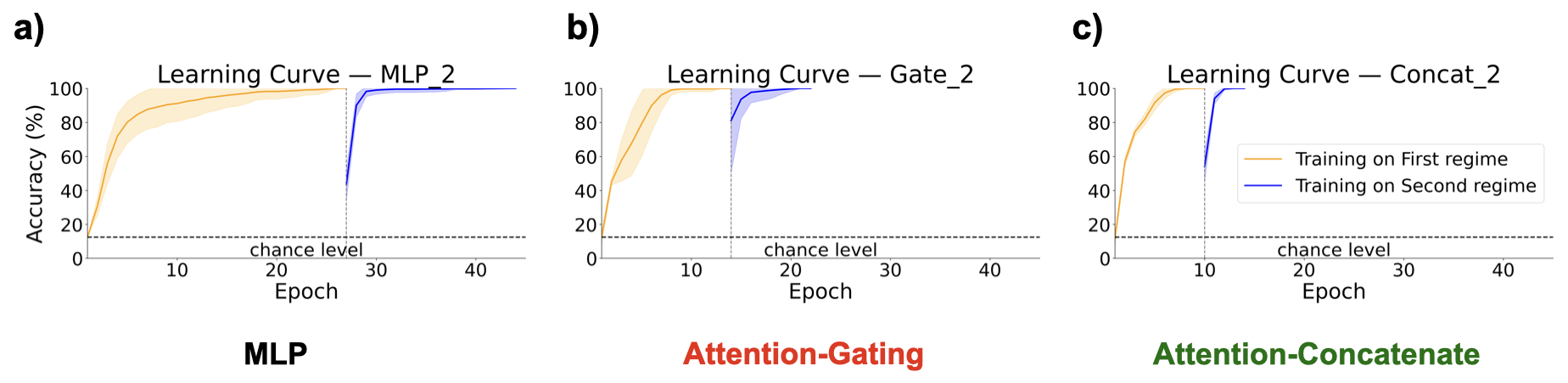}
\caption{\textbf{Learning curves in the Multi-4 Middle environment for 3 model types.}
Mean training accuracy (averaged over 50 runs) across two learning steps for (a) MLP\_2, (b) Gate\_2, and (c) Concat\_2. Orange curves show accuracy during learning step 1 (training on first-regime tasks), and blue curves show accuracy during learning step 2 (training on second-regime tasks). Shaded regions indicate ±1 standard deviation across runs. The vertical dashed line marks the transition between learning steps.}
\label{fig5}
\end{figure}

Fig~\ref{fig5} shows learning curves in the Multi-4 Middle environment for three model types (MLP, Attention-Gating, and Attention-Concatenate). All models reach near-ceiling accuracy ($\approx 100\%$) by the end of each learning step. Differences at the start of learning step 2 reflect variation in generalization to the second regime prior to training. In the following, we focus on generalization and stability performance.

\subsubsection*{Environmental Richness Enhances Generalization and Stability, Especially in Attention-Based Models}

\begin{figure}[!h]
\includegraphics[width=\textwidth]{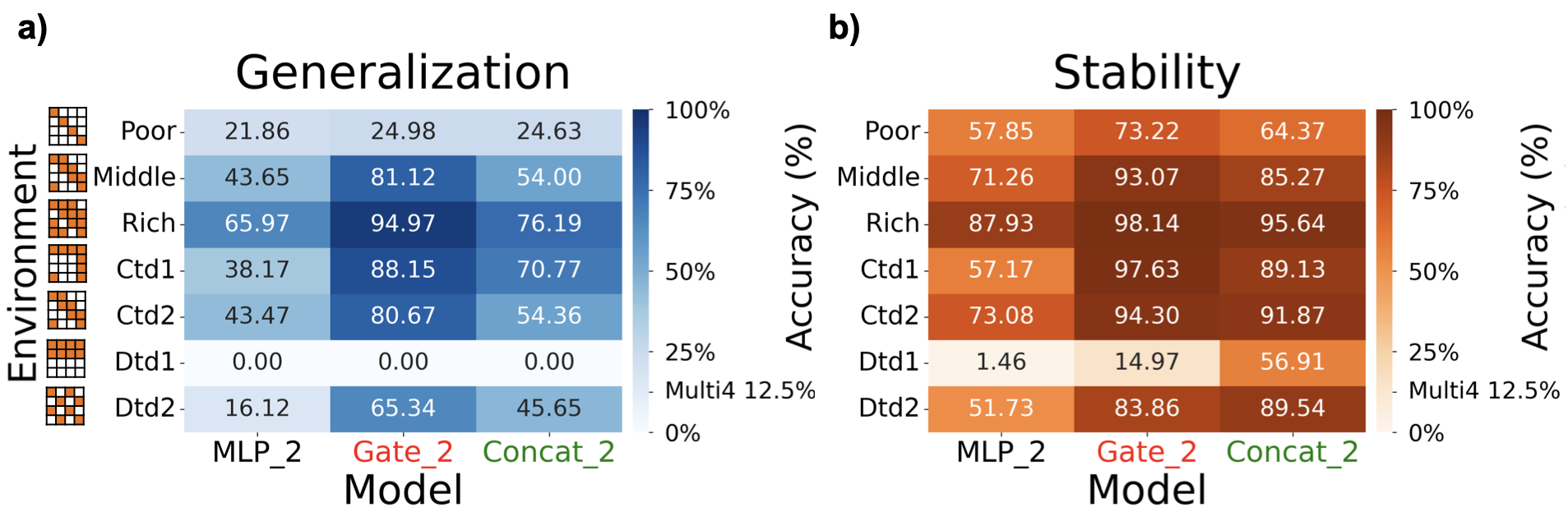}
\caption{\textbf{Effects of environmental richness and first-regime connectivity on generalization and stability in Multi-4.}
Mean accuracy (averaged over 50 runs) for MLP\_2, Gate\_2, and Concat\_2. (a) Generalization and (b) stability in Multi-4. Richness and connectivity improve both metrics for all models (connected $>$ disconnected), but attention-based models consistently outperform the MLP model. In the middle environment, connected regimes allow attention-based models to approach rich-environment performance, whereas the MLP model remains below rich levels and shows greater forgetting.  The chance level for each environment is indicated on the color bar. (Ctd = Connected; Dtd = Disconnected; middle environment.) 
}
\label{fig6}
\end{figure}

For Multi-4, to analyse the impact of environment richness on the model performance, we evaluate all models’ generalization and stability across poor, middle, and rich environments  (Fig~\ref{fig6}ab, rows 1-3). Across all environments, attention-based models significantly outperform the MLP model in generalization (Fig~\ref{fig6}a, rows 1–3). Although both model types benefit from increasing environmental richness, the performance gap remains substantial, particularly in the middle (MLP\_2: 43.65\%; attention-based models: 54.00--81.12\%) and rich environments (MLPs: 65.97\%; attention-based models: 76.19--94.97\%). These results suggest that attention-based models can leverage environmental richness more effectively to generalize to novel tasks.

Stability increases with environmental richness for all models, yet attention-based models consistently outperform the MLP model on each level of richness. The MLP model shows clear evidence of catastrophic forgetting, particularly in the poor and middle environments, where stability remains comparatively low (Multi-4 Poor: 57.85\%; Middle: 71.26\%). Attention-based models retain substantially more (Multi-4 Poor: 64.37--73.22\%; Middle: 85.27--93.07\%). Although richness improves stability for all models in Multi-4 Rich, attention-based models still achieve markedly higher (near-ceiling) robustness (MLPs: 87.93\%; attention-based: 95.64--98.14\%), indicating strong retention and near-complete preservation of prior knowledge. 

Overall, these findings indicate that attention-based models not only generalize better, but also reduce task interference (retain stability) more effectively than the MLP model.

\begin{figure}[!h]
\includegraphics[width=\textwidth]{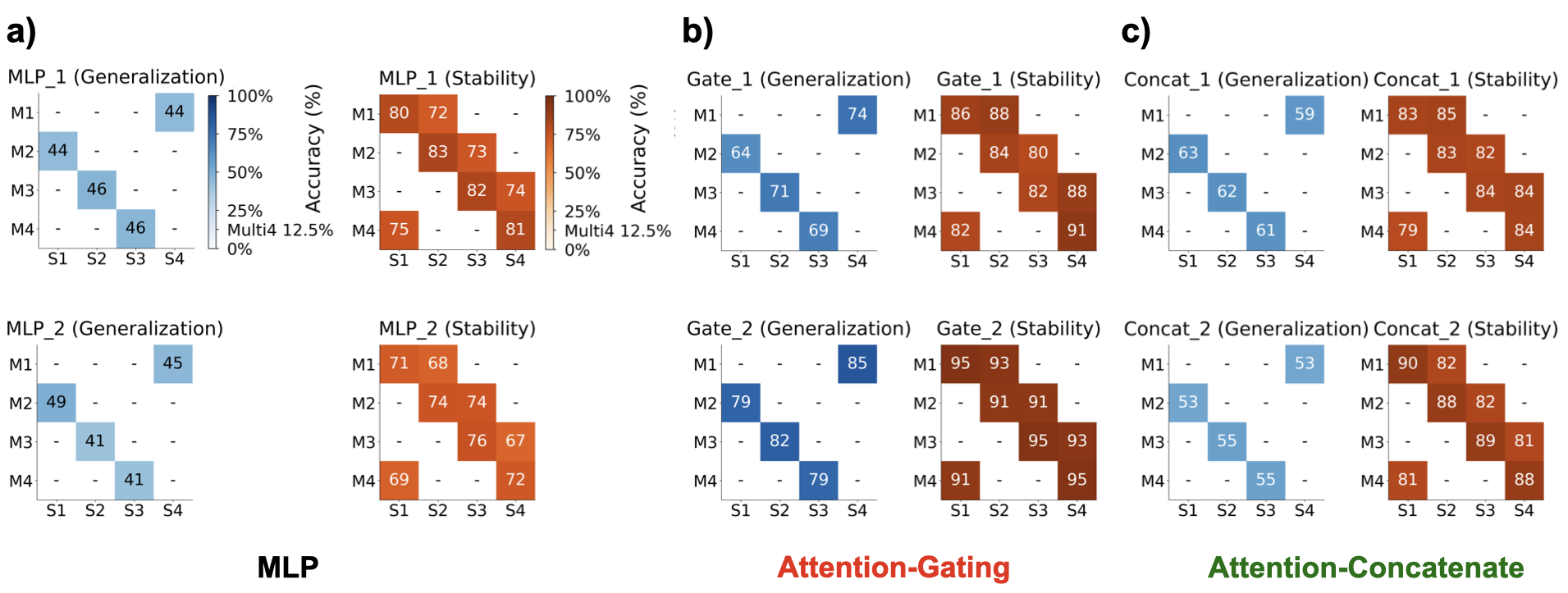}
\caption{\textbf{Task-wise performance in the Multi-4 Middle environment.}
(a) MLPs; (b) Attention–Gating models; (c) Attention–Concatenation models. Orange heatmaps show performance on the different tasks in the first regime, and blue heatmaps show performance on the different tasks in the second regime. Within each regime, accuracy varies only marginally across tasks, with no task on which any model is exceptionally strong or weak. Chance level (12.5\%) is indicated on the color legend.
}
\label{fig7}
\end{figure}

A possibility is that performance on novel tasks depends on the distance to the earlier (trained) tasks. To investigate this, we examined whether performance differs across tasks within the first and second regimes. Fig~\ref{fig7} presents task-wise accuracy for all models in each regime. In the Multi-4 Middle environment, generalization and stability accuracies are broadly similar across tasks, irrespective of their distance to trained tasks with only small variations within a given regime.

\subsubsection*{Generalization and stability in connected and disconnected environments}

We next investigate model performance across connected and disconnected first regimes in the middle environment, at a fixed level of richness (see Fig~\ref{fig6}a,b, rows 4-7). We examine the four first regimes (rows 4-5 are connected regime, rows 6-7 are disconnected). Connectedness influences the MLP model generalization (Fig~\ref{fig6}a, rows 4-7): performance is higher in connected than disconnected regimes (38.17--43.47\% vs 0--16.12\%). Attention-based models show the same qualitative pattern as the MLP model, performing better in connected than disconnected regimes (54.36--88.15\% vs 0--65.34\%), indicating that they can exploit task-structure connectivity far more effectively than the MLP baseline. 

A similar trend is evident for stability (Fig~\ref{fig6}b, rows 4–7): all models are more stable in connected than in disconnected regimes (MLP\_2: 57.17-73.08\% vs 1.46-51.73\%; Attention-based models: 89.13--97.63\% vs 14.97--89.54\%). However, the magnitude of this benefit differs sharply by architecture. For the MLP model, stability improves with connectedness but still remains relatively low, indicating continued vulnerability to catastrophic forgetting even in the connected regimes. In contrast, attention-based models capitalise much more strongly on connectedness, maintaining high stability in connected regimes.

Together, these results show that all models perform better in connected than in disconnected regimes; however, attention-based models consistently achieve higher generalization and stability than the MLP model, and are able to exploit task-structure connectivity to sustain high performance even in less rich environments.

\subsubsection*{Generalization and stability as a function of connectivity }

\begin{figure}[!h]
\includegraphics[width=\textwidth]{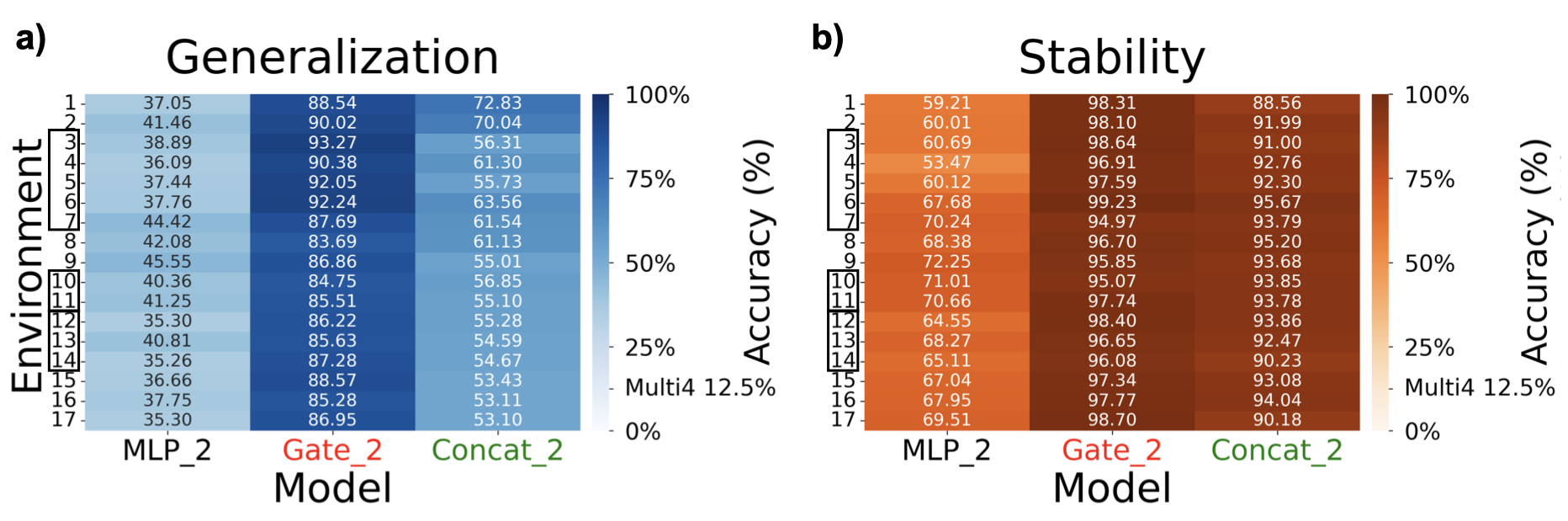}
\caption{\textbf{Accuracy across the 17 unique connected first-regime variants in the middle environment.}
Mean accuracy (averaged over 50 runs) for MLP\_2, Gate\_2, and Concat\_2. (a) generalization accuracy; (b) stability accuracy. Attention-based models outperform the MLP model on generalization and stability under all connected first regimes.  Regimes are ordered by increasing ASPL (higher to lower connectivity). Black boxes on the y-axis indicate identical ASPL (i.e., matched connectivity).
}
\label{fig8}
\end{figure}

To further investigate how, within the connected regimes, first regime task connectivity structure affects model performance, generalization and stability of the MLP and attention-based models are analysed with richness fixed but varying the connectivity of the first regime. From all possible task combinations, there are just 17 unique regimes in the Multi-4 middle environment (see Methods). 

Fig~\ref{fig8} summarizes accuracy across the 17 connected first-regime variants. From rows 1–17, ASPL and LSPL increase, indicating progressively weaker connectivity across rows. In Fig~\ref{fig8}a (generalization), MLP performance shows little systematic variation across regimes and remains low (35.30--45.55\%). By contrast, attention-based models consistently outperform MLPs, with the strongest results from Attention–Gating (83.69--93.27\%) and robust performance from Attention–Concatenation (53.10--72.83\%). In Fig~\ref{fig8}b (stability), the MLP model again exhibits comparatively low stability (53.47--72.25\%), consistent with catastrophic interference. Attention-based models remain superior across all 17 regimes, achieving high and often near-ceiling stability (90.18--99.23\%).

\begin{figure}[!h]
\includegraphics[width=\textwidth]{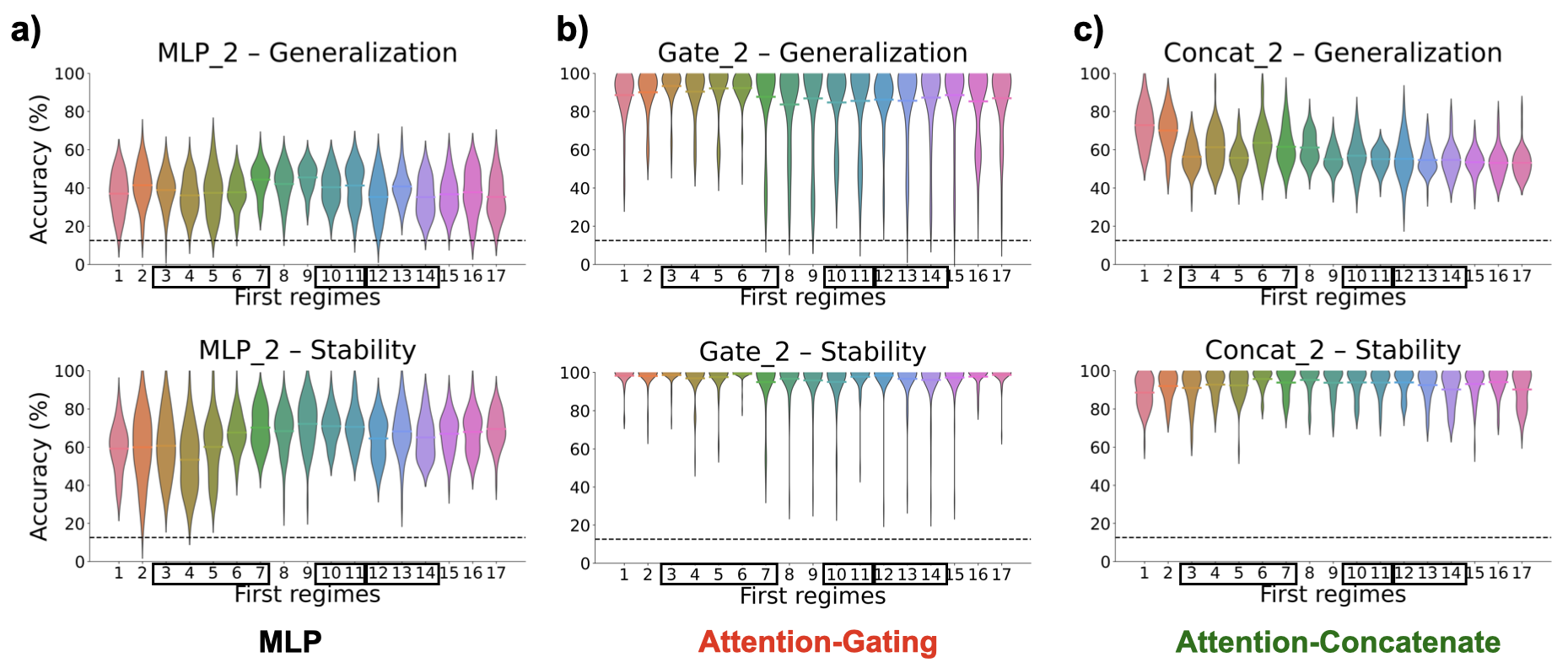}
\caption{\textbf{Violin plots of performance across 17 unique connected first-regime variants in the middle environment.}
Distributions of generalization and stability across 50 runs for (a) MLP\_2, (b) Gate\_2, and (c) Concat\_2. Regimes are ordered by increasing ASPL (lower connectivity), and black boxes group regimes with identical ASPL values. Each violin plot summarizes the distribution across runs, with the horizontal bar indicating the mean. Chance level is indicated by a dashed line in each plot.
}
\label{fig9}
\end{figure}

To visualize the variability in performance results, Fig~\ref{fig9} presents violin plots of generalization and stability across all 17 connected first-regime variants in the middle environment. In generalization (Fig~\ref{fig9}, top), MLP\_2 performs poorly and varies substantially across regimes, with no consistent dependence on connectivity. Gate\_2 remains near-ceiling, showing only a slight reduction as connectivity decreases, and Concat\_2 displays the same overall trend at a lower level while still reliably outperforming MLP\_2. In stability (Fig~\ref{fig9}, bottom), MLP\_2 stays relatively low with only a modest improvement across regimes, whereas Gate\_2 and Concat\_2 maintain consistently high, near-perfect stability throughout.

\begin{figure}[!h]
\includegraphics[width=\textwidth]{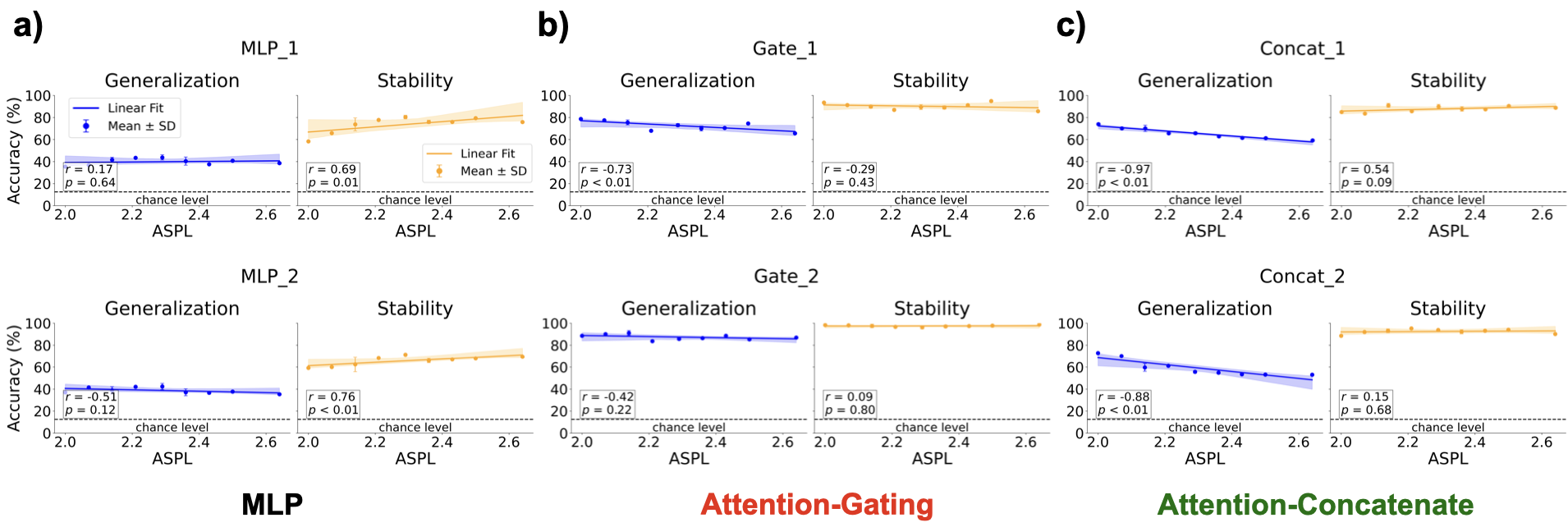}
\caption{\textbf{Linear regression of model performance vs. first-regime connectivity (ASPL) across 17 unique connected regimes.}
(a) MLPs show no significant relationship between ASPL and generalization accuracy, but exhibit a positive association between ASPL and stability. (b–c) For attention-based models, Gate\_1 and Attention–Concatenate models show negative correlations between ASPL and generalization accuracy. By contrast, Gate\_2 shows no significant correlation with ASPL for generalization, most plausibly due to a ceiling effect; the same ceiling effect likely explains the absence of a significant relationship for stability.
}
\label{fig10}
\end{figure}

Gate\_2 shows no significant correlation with ASPL for generalization, most plausibly due to a ceiling effect; the same ceiling effect likely explains the absence of a significant relationship for stability.

\begin{figure}[!h]
\includegraphics[width=\textwidth]{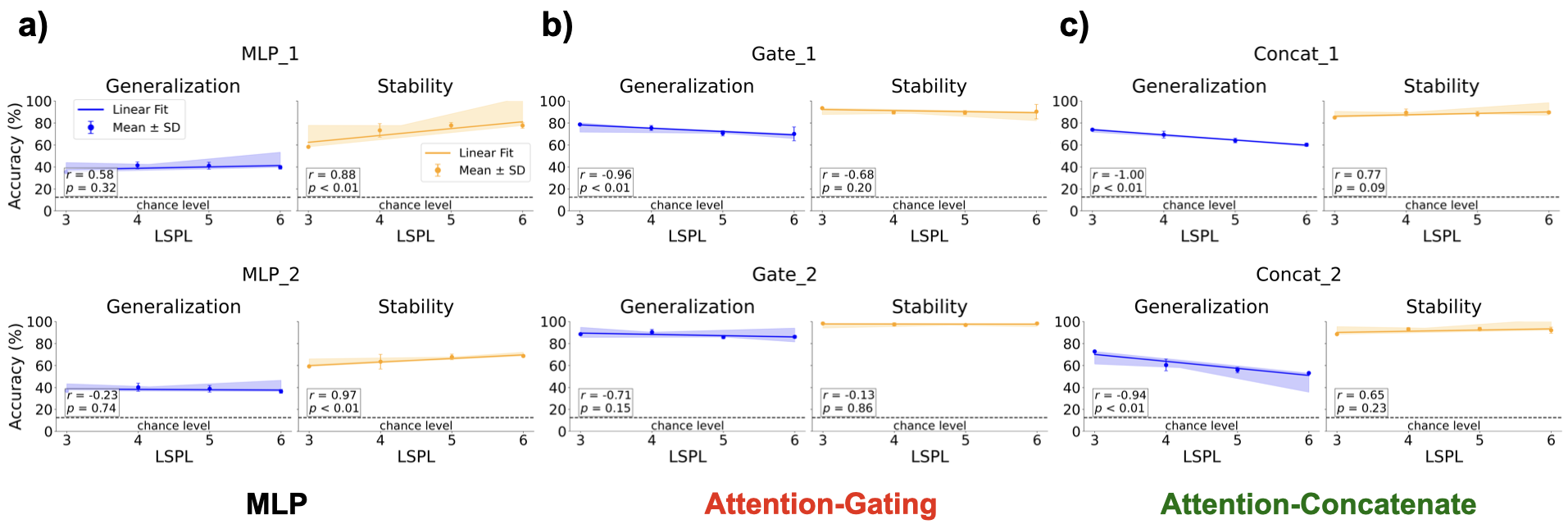}
\caption{\textbf{Linear regression of model performance vs. first-regime connectivity (LSPL) across 17 unique connected regimes.}
(a) MLPs show no significant relationship between LSPL and generalization accuracy, but exhibit a positive association between LSPL and stability. (b–c) For attention-based models, Gate\_1 and Attention–Concatenate models show negative correlations between LSPL and generalization accuracy. By contrast, Gate\_2 shows no significant correlation with LSPL for generalization, most plausibly due to a ceiling effect; the same ceiling effect likely explains the absence of a significant relationship for stability.
}
\label{fig11}
\end{figure}

We then examine the correlation between first-regime connectivity (measured by ASPL and LSPL) and model performance (generalization and stability) across the 17 unique connected first regimes (Figs~\ref{fig10} and ~\ref{fig11}). Regarding generalization, MLPs show no reliable relationship with connectivity. Accuracy is not consistently associated with either ASPL (MLP\_1: $r = 0.17$, $p = 0.64$; MLP\_2: $r = -0.51$, $p = 0.12$; Fig~\ref{fig10}a, top) or LSPL (MLP\_1: $r = 0.58$, $p = 0.32$; MLP\_2: $r = -0.23$, $p = 0.74$; Fig~\ref{fig11}a, top), suggesting that MLPs do not systematically exploit regime structure to support generalization. In contrast, attention-based models show a clear connectivity dependence: generalization accuracy is negatively correlated with both ASPL (Gate\_1: $r = -0.73$, $p < 0.01$; Concat\_1: $r = -0.97$, $p < 0.01$; Concat\_2: $r = -0.88$, $p < 0.01$; Fig~\ref{fig10}b,c, top) and LSPL (Gate\_1: $r = -0.96$, $p < 0.01$; Concat\_1: $r = -1.00$, $p < 0.01$; Concat\_2: $r = -0.94$, $p < 0.01$; Fig~\ref{fig11}b,c, top). In other words, more connected regimes (shorter ASPL/LSPL) support better generalization in attention-based models. The absence of a strong correlation for Gate\_2 (ASPL: $r = -0.42$, $p = 0.22$; LSPL: $r = -0.71$, $p = 0.15$) is most plausibly explained by a ceiling effect (Figs~\ref{fig10} and ~\ref{fig11}, Gate\_2 panels).

Regarding stability, MLPs show a clear positive relationship with both connectivity metrics: stability increases with ASPL (MLP\_1: $r = 0.69$, $p = 0.01$; MLP\_2: $r = 0.76$, $p < 0.01$; Fig~\ref{fig10}a, bottom) and with LSPL (MLP\_1: $r = 0.88$, $p < 0.01$; MLP\_2: $r = 0.97$, $p < 0.01$; Fig~\ref{fig11}a, bottom). This suggests that MLPs retain previously learned tasks more effectively when the first-regime structure is less connected (i.e., longer path lengths). By contrast, Attention–Gating and Attention–Concatenation models show no significant relationship between ASPL/LSPL and stability (Figs~\ref{fig10} and ~\ref{fig11}b,c, bottom), most likely because stability is already near-ceiling across regimes, leaving little room for a detectable connectivity effect.

\subsubsection*{Cue sensitivity }

\begin{figure}[!h]
\includegraphics[width=\textwidth]{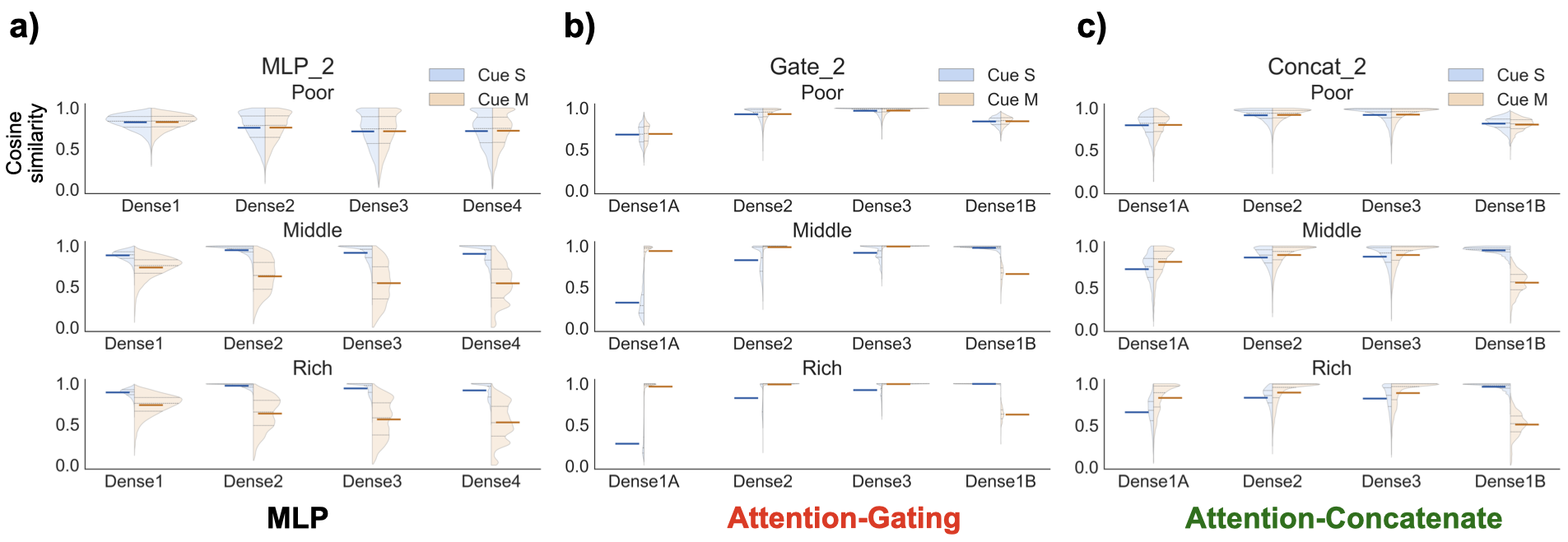}
\caption{\textbf{Cue sensitivity across poor, middle, and rich environments.}
Plots show mean cosine similarity of hidden representations under a single-cue change (averaged over 50 runs) for (a) MLP\_2, (b) Gate\_2, and (c) Concat\_2. Cue1 (blue) denotes the sensory cue and Cue2 (orange) the motor cue. Lower cosine similarity indicates  greater sensitivity to that cue change.
}
\label{fig12}
\end{figure}

To clarify why the gating and concatenation architectures outperform the MLP, we test the hypothesis that better stability and generalization stemmed from more compositionality in task representations in attention models (Fig~\ref{fig12}).  For this purpose, we examine layer-wise cue sensitivity. We quantified cue sensitivity as the cosine similarity between hidden-layer representations for trained vs. untrained tasks when only one cue was changed—either the sensory cue (blue) or the motor cue (orange). If a layer has low cosine similarity, this indicates that it is sensitive to the relevant (sensory or motor) dimension. Instead, if it has high cosine similarity, it is insensitive to changes in that dimension. These layers are labeled as Dense1, Dense2, etc, see Methods (Fig~\ref{fig14}) for details. We also assessed how this representational organisation varies with environmental richness.

Across environments, MLP\_2 exhibits only low-to-moderate sensitivity to sensory cue changes (Fig~\ref{fig12}a). As richness increases (from poor to middle/rich), MLP\_2 becomes more sensitive to motor cue changes, but this effect is diffuse rather than layer-specific. No layer shows a clearly “protected” representation that is selectively sensitive to one dimension while remaining insensitive to the other; instead, sensory and motor information remain largely entangled throughout the hierarchy.

In contrast, the attention-based models show increasingly structured cue sensitivity as richness increases (Fig~\ref{fig12}b,c). In the poor environment, both Gate\_2 and Concat\_2 exhibit relatively weak, diffuse cue sensitivity across layers. However, in the middle and rich environments, a clear layer-wise organisation emerges. In Gate\_2, Dense1A is selectively sensitive to sensory cue changes and Dense1B to motor cue changes, while later layers (Dense2/Dense3) show little sensitivity to either cue, suggesting that cue information is extracted and structured early and then passed downstream in a more stable form. Concat\_2 follows the same pattern, albeit with weaker sensory cue sensitivity in Dense1A, while motor-cue sensitivity remains pronounced in Dense1B and downstream layers remain high.

\begin{figure}[!h]
\includegraphics[width=\textwidth]{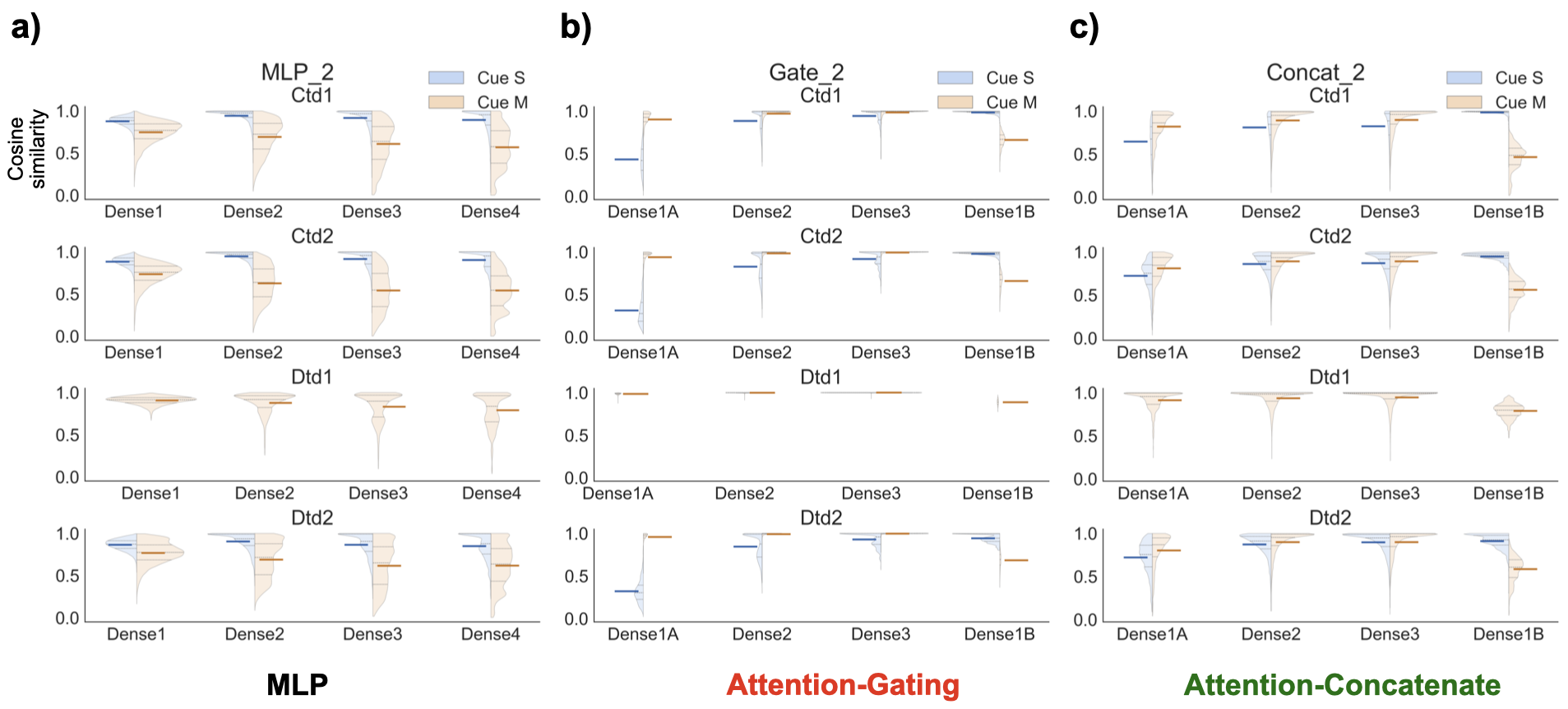}
\caption{\textbf{Cue sensitivity in the middle environment under connected (Ctd) and disconnected (Dtd) first regimes.}
Plots show mean cosine similarity of hidden representations under a single-cue change (averaged over 50 runs) for (a) MLP\_2, (b) Gate\_2, and (c) Concat\_2. Cue1 (blue) denotes the sensory cue and Cue2 (orange) the motor cue. Lower cosine similarity indicates greater sensitivity to that cue change. Ctd/Dtd correspond to the regimes in Figure 4; Dtd\_1 includes only sensory-cue changes.
}
\label{fig13}
\end{figure}

To test whether first-regime connectedness changes representational cue sensitivity, we compared connected and disconnected regimes in the middle environment (connected regimes: row 1-2; disconnected regimes: row 3-4, Fig~\ref{fig13}). Across all three architectures, cue sensitivity profiles are highly similar between the two conditions, indicating that connectivity has little to no effect on how cues are represented within the network. In both connected and disconnected regimes, MLP\_2 shows low sensitivity to sensory-cue changes across layers and moderate sensitivity to motor-cue changes (Fig~\ref{fig13}a) . Gate\_2 again exhibits cue-selective sensitivity concentrated in Dense1A (sensory) and Dense1B (motor), with minimal sensitivity in later layers, and Concat\_2 shows the same overall pattern, albeit with weaker sensory-cue sensitivity in Dense1A (Fig~\ref{fig13}b,c).

Overall, the layer-wise cue sensitivity analyses suggest that the superior stability and generalization of the attention-based architectures (Gate\_2 and Concat\_2) arises from more structured, cue-separable representations that become increasingly pronounced with environmental richness. Whereas MLP\_2 maintains entangled sensory and motor information across layers—with only diffuse, low-to-moderate cue sensitivity—Gate\_2 and Concat\_2 develop a clear early division of labour in the middle and rich environments. Dense1A is selectively sensitive to sensory cues and Dense1B to motor cues, while downstream layers become largely insensitive to both, consistent with early cue extraction followed by more stable, cue-insensitive representations. This cue-selective organisation is strongest in Gate\_2, somewhat weaker for sensory cues in Concat\_2, and is largely unchanged by first-regime connectivity in the middle environment, indicating that richness—not connectivity—primarily drives the emergence of compositional cue representations in these models.

\section*{Discussion}

Biological and artificial agents operating in dynamic environments must learn and switch between multiple tasks, creating opportunities to reuse shared structure but also risks of cross-task interference. In this work, we examined how performance in multi-task learning depends jointly on model architecture (MLPs vs attention-based models) and task environment structure by manipulating environmental richness (diversity of components) and connectivity (overlap of components across regimes). We found that increasing richness consistently improved both generalization and stability across model classes, with attention-based models outperforming MLPs throughout, consistent with the emergence of more cue-insensitive representations that can be reused across regimes without being overwritten. Connectivity further amplified generalization, especially in attention models. In these models, generalization increased approximately linearly with regime connectivity, while stability reached a ceiling in connected regimes. In contrast, MLP stability decreased with increasing connectivity, suggesting that overlapping components across tasks induce interference when the architecture lacks mechanisms for selective reuse. Notably, adding a bottleneck to attention-based models did not yield additional efficiency benefits: bottlenecked variants did not achieve better generalization–stability performance than their non-bottleneck counterparts. Together, these results show that cognitive flexibility depends on an interaction between task structure and model architectural capacity for efficiently exploiting representational overlap.

Across all model classes, we found that richer environments improved both generalization and stability, leading to higher cognitive flexibility. This fits with earlier work showing that multi-task learning can help learners discover compositional structure when the environment contains enough variety~\cite{johnston2023abstract,yang_task_2019,dorrell2025rangeindependencedrivesmodularity}. In a rich environment, agents observe many different combinations of the same underlying components. This makes it easier to identify which components are reusable, and therefore to generalize to new tasks that recombine familiar parts in novel ways.

Notably, the generalization–stability trade-off diminished as richness increased. This pattern follows from our evaluation setting, which targets compositional generalization: test tasks recombine components that were already present during training. With greater richness, components appear more frequently and in more diverse contexts, enabling more reliable component-based representations. Because these same representations remain useful for earlier tasks, learning new recombinations places less pressure on remapping or overwriting prior knowledge, improving stability alongside generalization.

However, richness does not only increase diversity—it also tends to increase connectivity. When more component dimensions are available and actively used, tasks are more likely to share components, creating more overlap between tasks and regimes. This means that the benefits we observe in rich environments could partly reflect the fact that the task space becomes more connected, providing more opportunities for knowledge reuse. To disentangle these factors, we therefore explicitly manipulated regime connectivity and tested how connectivity itself shapes generalization and stability across architectures.

Overall, connectedness mattered for performance: all models performed better in connected regimes than in disconnected regimes, indicating that shared components support learning and transfer. However, models differed in how they responded to connectivity, that is, to how strongly regimes were connected within the connected condition.
For MLPs, generalization was higher in connected than in disconnected regimes, but varied little as connectivity increased (as measured with ASPL or LSPL). This suggests that MLPs can benefit from the presence of shared structure, yet only to a limited extent. Component overlap provides some transferable information, but MLPs do not appear to make increasing use of stronger connectivity, such as shorter paths between shared components. At the same time, MLP stability decreased as connectivity increased, consistent with the idea that greater component sharing produces stronger interference in architectures without selective routing. When connectivity is lower, tasks share fewer components and can therefore be stored more independently, helping preserve performance on earlier tasks.
In contrast, attention-based models benefited not only from connectedness but also strongly from connectivity. Their generalization increased approximately linearly with connectivity, indicating that attention mechanisms are better able to exploit increasingly connected regimes for compositional transfer. Moreover, attention models reached ceiling-level stability in connected regimes, suggesting that once tasks are sufficiently connected, attention can reuse shared structure without overwriting prior knowledge. Together, these results suggest that connectivity becomes advantageous when the architecture can selectively control how shared representations are reused.
These findings also highlight an important interaction between richness, connectedness, and connectivity. Although richer environments improved performance across architectures, connectedness and connectivity provided an additional and previously underexplored source of structure. More generally, the mere presence of connectedness already confers an advantage, because even relatively weakly connected regimes support better performance than fully disconnected ones. Notably, even in environments with limited richness, highly connected regimes allowed attention-based models to approach the performance observed in richer settings. This suggests that greater connectivity can partially compensate for limited diversity by creating more opportunities for structured reuse of shared components across tasks.

This dependence on environmental structure can be understood from the perspective of the No Free Lunch theorem, which states that no learning algorithm can be universally (i.e., over all data) better: If some algorithm is better than another one on some data, there must be other data in which the other algorithm performs better~\cite{wolpert2002no}. Here, the attention-based models could profit from task structure, but this must mean that there are other task structures in which the MLPs fare better. It just so happens that modular task structure also characterizes our own world, making the attention-based models plausible as a cognitive architecture. 

These architecture-specific effects highlight that task performance is shaped not only by the model architecture, but by the global organization of tasks in the environment. Prior work has largely examined task structure through task similarity and has shown that partially overlapping tasks can be especially prone to interference~\cite{holton2026humans,saxe19}. Here, we extend this view by representing task environments as component-sharing graphs and systematically manipulating connectivity as a global property of the task space. This graph-based approach provides a principled way to quantify task structure beyond pairwise similarity and reveals that well-structured task relationships can support long-term retention by encouraging stable reuse of shared components rather than destructive remapping—especially for architectures, such as attention-based models, that can selectively exploit these relationships.

To better understand why attention-based models benefit more from connectivity, it is useful to consider how attention changes the way task information is processed. We implemented two methods for attention. The first one (gating) obtained inspiration from the cognitive neuroscience literature where attention is standardly applied via multiplicative gating~\cite{cohen1990control,VERBEKE2022256}. Attention-Gating models selectively filter stimulus information, allowing them to focus on relevant task features. The second one (concatenation) was inspired by AI literature, where transformer architectures~\cite{vaswani_attention_2017} implement binding by concatenating vectors A and B; this is accomplished by adding vector A representations to the vector B processing stream~\cite{elhage2021mathematical}. Irrespective of how it is implemented, attention-based models can decompose tasks and disentangle task information, which helps to generalize to novel tasks, and avoid mixing old and new task mappings. In contrast, MLPs process cues and stimuli together without task-specific filtering. This limits their ability to adapt to new tasks while retaining knowledge of previously learned tasks, especially in environments with overlapping or complex cue-task mappings.

To connect these architectural differences to underlying representational mechanisms, we used cue sensitivity under single-cue perturbations as an operational measure of disentanglement. If changing one cue causes the model to update only the cue-relevant sub-representation while leaving other cue representations and the shared task structure largely intact, this suggests that the task has been decomposed into separable components—that is, that the model has learned a disentangled feature representation. Cue sensitivity therefore reflects not only component separation, but also the selective deployment of the appropriate component: localized, cue-specific sensitivity is exactly what we would expect if the model is effectively attending to the relevant part of its representation rather than globally reorganizing what it encodes. Consistent with this interpretation, attention-based models exhibit more structured, layer-wise cue sensitivity than MLPs, whereas MLPs process cues and stimuli in a more uniformly mixed manner, limiting their ability to adapt to new tasks while preserving previously learned mappings, especially when cue–task structure is overlapping or complex.
Importantly, environmental richness strengthens this process in both model classes. Richer training regimes provide more varied evidence about what is shared and what is cue-specific, helping models learn cleaner decompositions and more reliable cue-guided selection. However, disentanglement, as captured by cue sensitivity, is not sufficient to explain robust compositional generalization~\cite{montero2021the}. In our results, cue sensitivity varied little with connectedness, even though performance clearly improved in connected regimes and often increased further with greater connectivity. This suggests that prior work misses an additional factor: task connectivity. Whereas richness primarily supports the learning of decomposed representations, connectedness and connectivity determine whether those representations can be usefully reused across regimes. Connectedness and connectivity increase the functional value of learned components by creating more opportunities for structured reuse. This helps explain the gap left by earlier accounts that focused on disentanglement alone.

These results also suggest a broader perspective on continual learning. Much of the literature treats catastrophic interference as a problem to be solved primarily through architectural or algorithmic interventions, such as regularization, replay or weight protection~\cite{kirkpatrick2017overcoming,mcclelland1995there,DBLP:journals/corr/ShinLKK17,grossberg_how_1980,VERBEKE2022256}. Our findings complement these approaches by highlighting a second, often overlooked factor: the structure of the task environment. In particular, our connectivity manipulation shows that interference is not only a property of the learner, but also of how tasks are related—overlap can either support transfer or induce forgetting depending on whether the model can selectively reuse shared representations.

From this viewpoint, several existing frameworks can be reinterpreted through the lens of connectivity and structured reuse. One example is curriculum learning, which studies how ordering tasks influences learning efficiency~\cite{10.1145/1553374.1553380}. While curriculum effects appear small in some standard supervised benchmarks ~\cite{wu_when_2021}, they can be substantial when tasks have strong dependencies ~\cite{matiisen}. This contrast fits naturally with our results: curricula should be most beneficial in high-connectivity environments, where later tasks can build on components learned earlier, and least beneficial when tasks are largely independent. This may also help explain why curricula are often effective in biological learning~\cite{dekker_curriculum_2022}, where tasks tend to implement structural dependencies.

Also in reinforcement learning (RL), environmental structure can directly support compositional transfer. When reward functions and transition dynamics combine modularly, agents can learn separable representations that generalize through recombination ~\cite{franklin_compositional_2018}, and humans show evidence of exploiting such modular structure ~\cite{tomov_multi-task_2021}. More broadly, graph-based views of task structure connect to RL work that analyses global task structure to identify bottleneck states (high-centrality states that serve as useful subgoals for learning~\cite{csimcsek2008skill}). Also the motivation for successor representations that are thought to occur in the hippocampus~\cite{stachenfeld2017hippocampus} starts from an analysis of the global task structure as formalized by the task’s graph Laplacian.  Such global analysis of task environment structure may offer a fruitful way to characterize how global task connectivity supports generalization and stability.

A primary limitation of the present work is that our task environments are relatively simple. While this simplicity is useful for isolating richness and connectivity, it remains to be tested whether the same principles scale to larger task spaces, noisier inputs, and settings with only partially observable task structure. A second limitation is that we operationalize connectivity as componential overlap across regimes; other forms of structure, such as hierarchical dependencies, temporal correlations, or causal relationships between task dimensions, may shape generalization and stability in different ways. Finally, although our invariance analyses provide evidence about representational organization, they capture only one aspect of internal structure and do not fully explain how representations are reused during learning.

Despite these limitations, our work shows that task environment structure (richness and connectivity) is an important component for explaining cognitive flexibility (generalization and stability) of neural networks when applied to multi-task learning. These findings open new avenues for investigating whether humans show similar sensitivity and take advantage of the structure of the environment. 

\section*{Materials and methods}

\subsection*{Multi-task structure}
As shown in Fig~\ref{fig1}a, Multi-2 is a $2 \times 2$ dimensional task structure with two sensory cues and two motor cues. Our task generalizes the task of Badre et al.~\cite{badre2010frontal}(which was hierarchical in the stimulus domain) to one that also imposes a hierarchy in the motor domain. The tasks are defined by unique combinations of sensory cues $(S_1, S_2)$ and motor cues $(M_1, M_2)$, resulting in a total of four possible tasks (cue combinations). A collection of tasks is called a regime. In Multi-2, the first regime consists of two tasks, and each task includes 5{,}000 trials. Every trial presents a single stimulus. Each stimulus is defined by one feature from each sensory dimension (e.g., color: red; shape: triangle). There are two sensory dimensions, and each has two possible values—for example, color can be red or blue, and shape can be triangle or circle. Combining these dimensions ($2 \times 2$) yields four distinct stimuli, representing all possible combinations of sensory features, and all four are included in the trials of each task. Fig~\ref{fig1}a illustrates an example of a first regime (orange tasks); and a second regime (blue tasks).
 
To investigate how multitask environments influence model performance, we vary the number of tasks in the first regime (i.e., regime used for learning step 1), to generate poor and rich environments. For this purpose, we used a Multi-3 structure (Fig~\ref{fig1}b). Multi-3 has 9 tasks (cue combinations: 3 sensory cues $\times$ 3 motor cues) and 8 unique stimuli ( 2 values in each sensory dimension). In the poor environment, the first regime includes three tasks, each associated with a distinct combination of sensory and motor dimensions (and their corresponding cues). These three tasks jointly cover all sensory and motor cues within the task structure. In the rich environment, the first regime includes six tasks, offering a more diverse set of training experience while still covering all dimensions of each modality. In both environments, the second regime consists of three tasks outside the first regime (blue regimes in Fig~\ref{fig1}b). We quantify richness as the proportion of first regime tasks relative to the total number of tasks in the task structure (33 and 66\% in poor and rich environments, respectively). 

\subsection*{Training and testing procedures}

The training and testing procedures are summarized in Fig~\ref{fig1}c. In Learning Step 1, models are trained on the first regime tasks (5{,}000 trials per task, 1 epoch). The model weights are frozen at the end of this step for further training. This is followed by a generalization test on the second regime tasks (without feedback), to evaluate model performance on unseen second regime tasks (generalization). In Learning Step 2, tasks from the second regime are learned with feedback. Finally, a stability test is conducted on the first regime (without feedback), to assess the model’s resistance to (catastrophic) forgetting (i.e., stability).

\subsection*{Task structure connectivity}

To further investigate how various environmental factors influence model performance, we also use the Multi-4 task structure (Fig~\ref{fig4}) offering increased complexity to explore these factors more thoroughly. The Multi-4 task structure comprises 16 possible cue combinations (i.e., tasks), formed by pairing 4 sensory cues with 4 motor cues. It also includes 16 unique stimuli, each corresponding to a distinct combination of sensory features (with 2 levels per sensory dimension, giving $2^4 = 16$ stimuli). 
 
Multi-4 has three levels of environmental richness: poor, middle, and rich (see Fig~\ref{fig4}a) The first regimes include 4, 8, and 12 tasks in poor, middle, and rich environments, with richness levels of 25\%, 50\%, and 75\% of all possible tasks, respectively. Again, the second regime consists of 4 tasks not included in the first regimes.

In the middle environment, eight tasks are selected from a total of 16 possible tasks to form the first regime. There are multiple ways to construct these regimes (see Fig~\ref{fig4}b), following the general rule that for $n$ dimensions, where a subset of T tasks are selected, there are $\binom{n^2}{T}$ possible regimes. For example, when $n = 4$ and $T = 8$ tasks are selected, this results in $\binom{16}{8}$ possible combinations. However, the specific sensory cues or motor cues within a regime does not carry meaningful differences. Changing the sensory cues or motor cues within a regime leads to an equivalent regime, as long as the connectivity structure is maintained. Similarly, swapping all sensory cues or all motor cues (as in a matrix transposition) within a regime yields an equivalent configuration. Therefore, regimes that can be generated by changing or transposing sensory and motor cues are treated as identical. To account for these equivalencies, unique regimes are defined as those that cannot be transformed into another regime through such operations. In the middle environment (where 8 training tasks are selected from 16 possible tasks), the number of unique regimes thus defined, equals 32~\cite{faradvzev1978constructive}. These unique regimes are sufficient to represent all possible regime configurations in the middle environment.
 
In addition to environment richness, we highlight connectedness between tasks as a key environmental factor that may influence model performance. We calculate connectivity in cue space, as shown in Fig~\ref{fig4}c, which illustrates the connectedness of regimes presented in Fig~\ref{fig4}b. Each cue is represented as a vertex in the graph, and each task corresponds to an edge connecting a sensory cue vertex $S_i$ to a motor cue vertex $M_j$. Since each task involves exactly one sensory and one motor dimension (cue), each regime can be represented as an undirected bipartite graph. 

We define a vertex $i$ and a vertex $j$ to be connected if there exists a path between them through a sequence of edges. For example, regimes that include tasks $(S_1, M_1)$ and $(S_2, M_1)$ create a path linking $S_1$ and $S_2$ through their shared connection to $M_1$. A regime is considered connected if there is a path between every pair of sensory and motor vertices. Conversely, a regime is disconnected if such a path does not exist.
 
In Fig~\ref{fig4}c the connectivity of regimes in Fig~\ref{fig4}b is illustrated. The first regime in Fig~\ref{fig4}b is connected, as seen in the corresponding graph in Fig~\ref{fig4}c, because there exists a path between every pair of sensory and motor vertices. The second regime in Fig~\ref{fig4}b is also connected. The third regime in Fig~\ref{fig4}b is disconnected, as shown in Fig~\ref{fig4}c, because there is no path connecting M$_3$ and M$_4$ to the other vertices. The fourth regime in Fig~\ref{fig4}b is also disconnected. Among the 32 unique regimes of the Multi-4 task structure, 17 are connected and 15 are disconnected.

\begin{table}[!ht]

\centering
\caption{\textbf{Average shortest path length (ASPL) and longest shortest path length (LSPL) of all unique regimes in the middle environment.}}
\begin{tabular}{lll} 
\hline
ASPL   & LSPL   & Number\\
\hline
2    & 3 & 1 \\
2.07 & 4 & 1 \\
2.14 & 4 & 5 \\
2.21 & 5 & 1 \\
2.29 & 4 & 1 \\
2.29 & 5 & 2 \\        
2.36 & 5 & 3 \\
2.43 & 5 & 1 \\
2.5  & 6 & 1 \\
2.64 & 6 & 1 \\
inf & inf & 15 \\
\hline
\end{tabular}
\begin{flushleft} Each row indicates the number of existing regimes in the middle environment with a given level of average shortest path length (ASPL) and longest shortest path length (LSPL).
\end{flushleft}
\label{table1}
\end{table}

To quantify connectivity, we apply methods from graph theory. The shortest path length measures the minimum number of edges required to travel between a pair of vertices. The average shortest path length (ASPL) represents the average shortest path length between all pairs of vertices, while the longest shortest path length (LSPL) denotes the maximum shortest path length among all pairs of vertices. In disconnected regimes, both ASPL and LSPL are defined to be infinite. Table~\ref{table1} shows the possible (ASPL, LSPL) combinations for Multi-4 and shows how many regimes exist for each (ASPL, LSPL) combination.

\subsection*{Models}

\begin{figure}[!h]
\includegraphics[width=\textwidth]{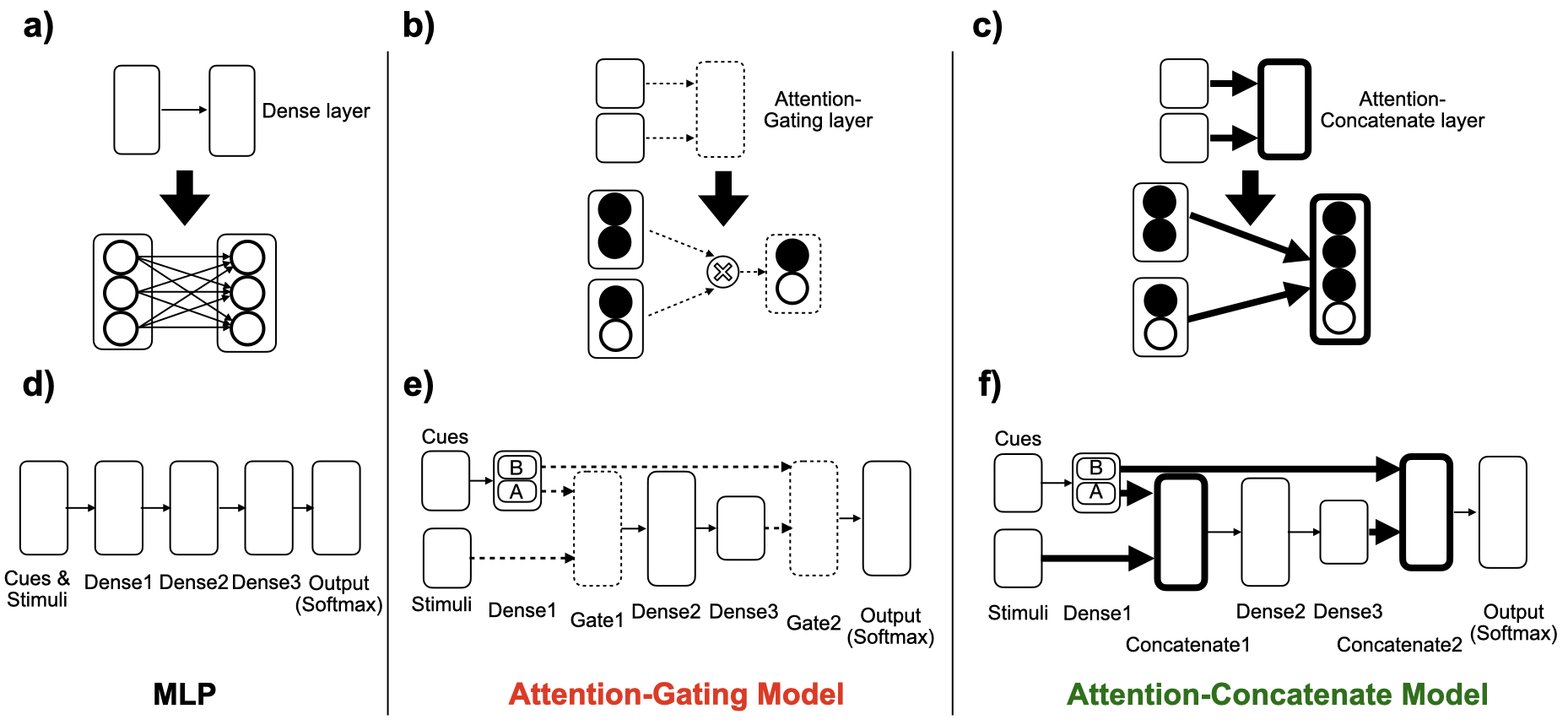}
\caption{\textbf{Model architectures (for Multi-4).}
(a) Dense layer in MLP. (b) Mechanism in Attention-Gating layer. (c) Mechanism in Attention-Concatenate layer. (d) MLP model processes cues and stimuli together with backpropagation. (e) Attention-Gating Model uses Attention-Gating layers (Gate1 and Gate2) to gate stimulus features guided by cue features. In the upper plot, filled circle means active, open circle inactive. (f) Attention-Concatenate Model concatenates cue and stimulus features to guide learning (Concatnate1 and Concatenate2). Dotted lines indicate attention-gating pathways. Full lines indicate attention-concatenate pathways.
}
\label{fig14}
\end{figure}

\subsubsection*{Multilayer Perceptron Models}
We used an MLP model as a baseline (Fig~\ref{fig14}d). The model takes cues and stimuli as input, both represented as binary vectors. For example, with two cues, the sensory cue input for color is represented as $[1, 1, 0, 0]$, and for shape as $[0, 0, 1, 1]$. With three cues, the color input becomes $[1, 1, 0, 0, 0, 0]$, and the shape input $[0, 0, 1, 1, 0, 0]$. The same encoding scheme is applied to the motor cues.

The stimulus input is coded similarly. For example, with two sensory features, $[1, 0, 1, 0]$ could represent (color: red; shape: triangle), and $[0, 1, 1, 0]$ (color: blue; shape: triangle). The total input size is the sum of all sensory cues, motor cues, and stimulus feature dimensions, calculated as $2 \times$(number of cues ($2n$)) + $2 \times$(number of stimulus sensory dimensions $n$)) = $6n$. The output is the predicted response, represented as a one-hot encoded vector of size $2n$.

\begin{table}[!ht]
\centering
\caption{\textbf{Number of learnable parameters in the hidden layers of each model.}}
\begin{tabular}{ccccccccccccc}
\toprule
 & \multicolumn{2}{c}{MLP\_1} 
 & \multicolumn{2}{c}{MLP\_2} 
 & \multicolumn{2}{c}{Gate\_1} 
 & \multicolumn{2}{c}{Gate\_2} 
 & \multicolumn{2}{c}{Concat\_1} 
 & \multicolumn{2}{c}{Concat\_2} \\
\cline{2-13}
 & $N_w$ & $N_l$
 & $N_w$ & $N_l$ 
 & $N_w$ & $N_l$ 
 & $N_w$ & $N_l$ 
 & $N_w$ & $N_l$ 
 & $N_w$ & $N_l$ \\
\toprule
Multi-2 
 & 14 & 2 
 & 18 & 3 
 & 10 & 2 
 & 12 & 3 
 & 10 & 2 
 & 12 & 3 \\
Multi-3 
 & 36 & 3 
 & 42 & 4 
 & 12 & 3 
 & 18 & 3 
 & 12 & 3 
 & 18 & 3 \\
Multi-4 
 & 48 & 3 
 & 56 & 4 
 & 26 & 3 
 & 36 & 3 
 & 26 & 3 
 & 36 & 3 \\
\bottomrule
\end{tabular}
\begin{flushleft} $N_w$ means the number of weights (learnable parameters). $N_l$ means the number of hidden layers.
\end{flushleft}
\label{table2}
\end{table}

The model uses a sigmoid activation function in each hidden layer and is connected with a softmax layer to the output. It is trained using backpropagation implemented in TensorFlow~\cite{abadi_tensorflow_2016}. We used sigmoid activation functions. Other activation functions (e.g., RELU) led to similar results. To optimize performance across environments, we vary model capacity slightly (Table~\ref{table2}). The architectures for Multi-3 and Multi-4 are the same and include one extra dense layer relative to Multi-2. Within each environment, we compare two network depths to determine the best-performing configuration—for instance, in Multi-4 we evaluate a three-hidden-layer MLP (MLP\_1) and a four-hidden-layer variant (MLP\_2). The only difference between MLP\_1 and MLP\_2 is the addition of this extra hidden layer. Table~\ref{table2} summarizes the number of hidden layers and total learnable parameters for each model.

Models were trained to perform all tasks within each regime using supervised learning. On each trial, the stimulus was provided to the stimulus-processing input stream and the corresponding task cue to the cue input stream (for MLPs, cues and stimuli were provided through the same input stream). Each task contributed 5000 trials, and trials from all tasks in the regime were randomly shuffled to form the training set. For each environment and model type, we ran 50 independent training runs with different random initializations. Training used a batch size of 32, the Adam optimizer, and categorical cross-entropy loss. Each run continued for multiple epochs and terminated early once the model achieved 100\% accuracy for four consecutive epochs. We report performance as the mean accuracy across runs. All weights were initialized using Xavier uniform initialization to ensure a well-scaled starting point for optimization.

\subsubsection*{Attention Models}
We next introduce Attention models, an extension of the baseline MLP designed to improve task decomposition and thus multidimensional task performance (Fig~\ref{fig14}e-f). These models utilize attention mechanisms to dynamically prioritize task-relevant information by assigning varying importance to different inputs, which can mimic the decomposition process in human cognitive behavior on complex tasks. The attention mechanism operates across multiple layers, where cues guide the model at different stages to process relevant information from the stimuli. The core idea of attention is to let the model selectively focus on important features, as guided by cues~\cite{hummos23,sommers25,VERBEKE2022256}, thereby enhancing its ability to generalize knowledge across tasks while maintaining stability by disentangling task-relevant information.
 
We operationalize attention by incorporating learned attention weights that modulate the inputs before they are processed by dense layers. To make robust conclusions about the role of attention, we test two attention-based architectures: Attention-Gating and Attention-Concatenate models (Fig~\ref{fig4}e-f). For each architecture, we evaluate two versions: one with a bottleneck dense layer (Gate\_1, Concat\_1, Fig~\ref{fig14}e-f) and one without (Gate\_2, Concat\_2). The bottleneck means that Dense2 contains just two units, creating a bottleneck that restricts how much information can flow through. We reason that such a bottleneck constraint forces the model to generate a simplified intermediate output, thereby improving interpretability and robustness while reducing the likelihood of overfitting.
 
In attention–based models (Attention-Gating and Attention-Concatenate models), the sensory and motor cues are first passed through Dense1, which learns a task-dependent transformation of the cue information during training (see Fig~\ref{fig14}e-f). The resulting representation is then split into two (learned) components (A and B), each providing conditioning information for a different processing stage. Specifically, A is fed to the first attention layer (Gate1 or Concat1) and B to the second (Gate2 or Concat2). Importantly, because Dense1 is a dense transformation, the model learns autonomously which cue features should modulate attention at each attention layer.

In the Attention-Gating model (Fig~\ref{fig14}e), information A from Dense1 is passed to the first Attention-Gating layer (Gate1), where it gates the stimulus inputs using a multiplicative gating mechanism. This mechanism selectively emphasizes task-relevant stimulus features while inhibiting irrelevant ones, mimicking how biological attention systems gate sensory information. Gated outputs are passed through Dense2 and Dense3, where task-relevant information is further processed. The output of Dense2 is then fed into the second Attention–Gating layer, which uses component B from Dense1 to further modulate and refine the intermediate representation, thereby promoting task-appropriate processing and response generation. 

In the Attention–Concatenate model (Fig~\ref{fig14}f), the cue pathway mirrors the Attention–Gating architecture: Dense1 produces two cue-derived components (A and B) that condition the first and second attention stages, respectively. However, rather than modulating stimulus representations via multiplicative gating, the model concatenates cue information (A and B) with the stimulus features produced by the dense layers. This joint representation allows cue signals to steer attention by indicating which aspects of the stimulus representation should be emphasised at each stage.

For model capacity, the attention-based architectures used in Multi-3 and Multi-4 are identical and include an additional dense layer inserted between the two attention layers relative to Multi-2. Within each environment, attention-based models share the same hidden-layer backbone as MLP\_1. For each attention mechanism, we consider two variants: a bottleneck version with a two-unit dense layer immediately before the second attention layer (Gate\_1 in Attention-Gating and Concat\_1 in Attention-Concatenate), and a non-bottleneck version without this compression (Gate\_2 and Concat\_2). The number of hidden layers and total learnable parameters for all models are reported in Table~\ref{table2}. The same training procedure was followed as for the MLP model, including input (one-hot) coding, training time, sigmoid activation, SoftMax layer, (Adam) optimizer, and weight initialization.

By systematically comparing different attention-based model architectures and mechanisms, we aim to identify how the combination of attention mechanisms and architectural constraints (e.g., a bottleneck) supports generalization and stability across multitask environments with varying levels of richness and connectivity.

\subsection*{Cue sensitivity analysis}

To quantify how cue changes within a task regime shape internal processing, we measured cue sensitivity at each hidden layer. Here, cue sensitivity reflects how well a layer preserves its representation of the stimulus (and the unchanged cue) when one cue dimension—sensory or motor—is altered between tasks.

Within each task regime, we formed matched task pairs between a trained task and an untrained task that differed in exactly one cue (sensory or motor) while holding the stimulus and the other cue fixed. For each trained–untrained pair, we presented an identical set of stimuli to the network and recorded the activation vector from every hidden layer. For instance, in a sensory-cue pair, the stimulus could be the same colored shape (e.g., a red triangle) and the motor cue could be fixed (e.g., “index finger”), while the sensory cue differs between the trained and untrained tasks (attend “color” vs. “shape”). We then feed the same stimulus set (red triangle, blue triangle, red square, blue square, \ldots) under each cue configuration and, for every stimulus, log the hidden-layer activation vector (i.e., the concatenated unit activations) from layer 1 through the final hidden layer. To reduce trial-level noise, we averaged activations over 30 repeated presentations for each stimulus–cue configuration.

Cue sensitivity for a given layer $L$ and task pair $(T_1,T_2)$ was then defined as the cosine similarity between the mean activation vectors elicited by the two tasks. Here, $h(T_1,L)$ and $h(T_2,L)$ denote the mean activation (across stimuli) of layer $L$ under tasks $A$ and $B$, respectively, where $A$ and $B$ differ in exactly one cue. Cue sensitivity was then computed as: 
\[
\mathrm{Sens}(L; T_1, T_2) = \cos\big(h(T_1, L),\, h(T_2, L)\big).
\]
We computed $\mathrm{Sens}(L; T_1, T_2)$ for all eligible trained–untrained task pairs and then averaged across stimuli and cue pairs to obtain a single sensitivity score per layer, separately for sensory-cue changes and motor-cue changes. Reported values are averages over 50 runs in which the model reached $98\%$ accuracy within the fixed number of training epochs.

Interpretation is based on the magnitude of cosine similarity: Higher cosine similarity indicates that the layer is highly insensitive to that cue change; the representational geometry is nearly unchanged when the task cue is altered. Lower cosine similarity indicates that the layer is strongly sensitive to that cue; the representational pattern changes substantially when the cue changes. Intermediate values reflect partial sensitivity.

\section*{Acknowledgments}
Xiaoyu K. Zhang is supported by Chinese Scholarship Council. We thank Steven Van Overberghe and Tom Lauwaerts for bringing our attention to the work on Isomorph-Free Exhaustive Generation.

\nolinenumbers

%
%

\bibliography{reference}
\section*{Supporting information}
In this appendix, we provide the full figures for all six models, including the models MLP\_1, Gate\_1 and Concat\_1 that were removed for brevity from the figures in the Results section of the main manuscript.

\includegraphics[width=\textwidth]{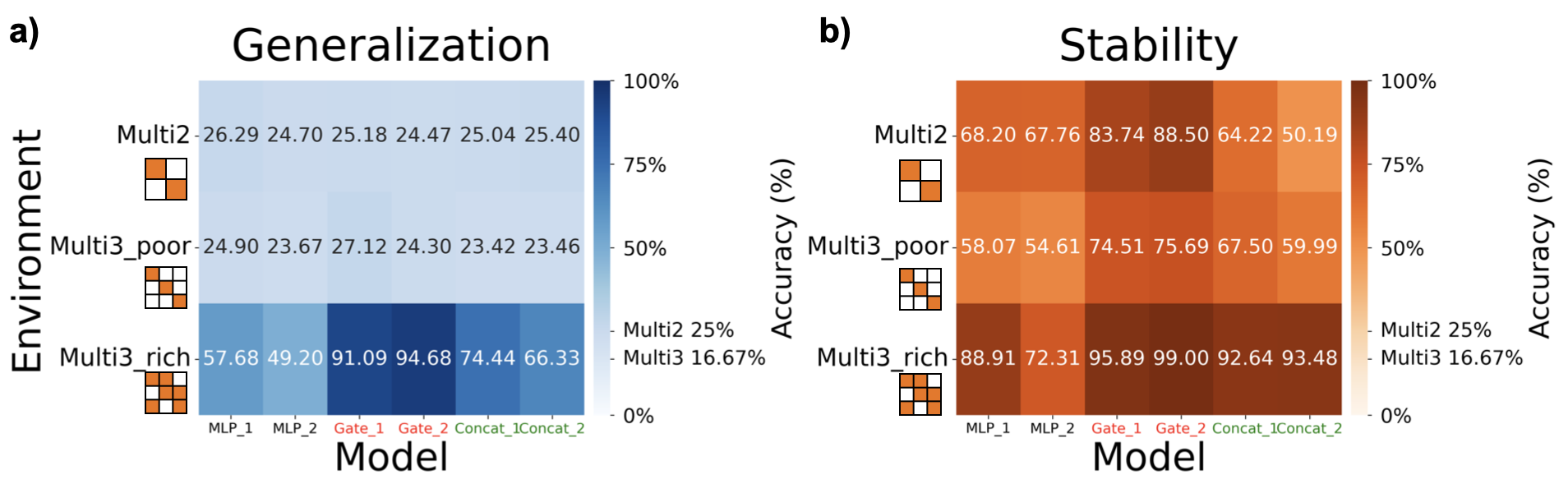}
\paragraph*{Fig. S1}
\label{S1_Fig}
\textbf{Performance of all models in generalization and stability across different environments (Multi-2, Multi-3 Poor, Multi-3 Rich).} (a) Generalization results show higher accuracy for attention-based models compared to MLPs, with near-perfect performance in the rich environment. (b) Stability results highlight the robustness of attention-based models in retaining prior knowledge, while MLPs exhibit catastrophic forgetting. All models perform better in the rich than poor environment in both generalization and stability. The chance level is indicated on each color bar.

\includegraphics[width=\textwidth]{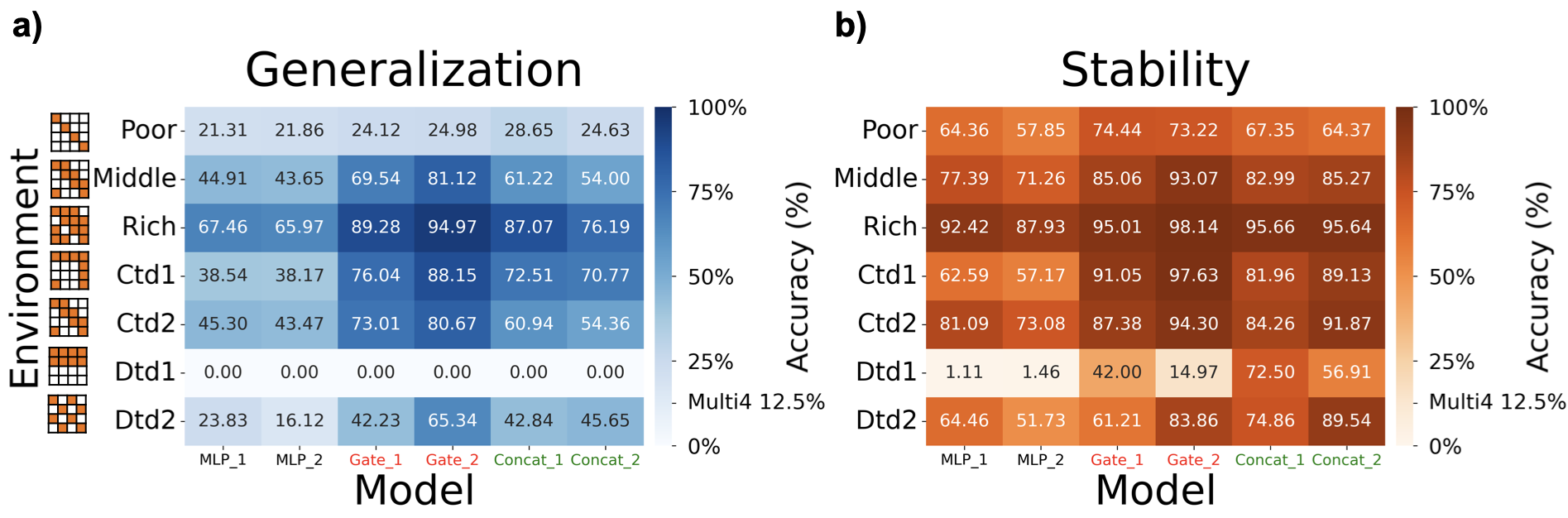}
\paragraph*{Fig. S2}
\label{S2_Fig}
\textbf{Effects of environmental richness and first-regime connectivity on generalization and stability for all models in Multi-4.} (a) Generalization and (b) stability in Multi-4. Richness and connectivity improve both metrics for all models (connected $>$ disconnected), but attention-based models consistently outperform MLPs. In the middle environment, connected regimes allow attention-based models to approach rich-environment performance, whereas MLPs remain below rich levels and show greater forgetting.  The chance level is indicated on each color bar. (Ctd = Connected; Dtd = Disconnected; middle environment.) 

\includegraphics[width=\textwidth]{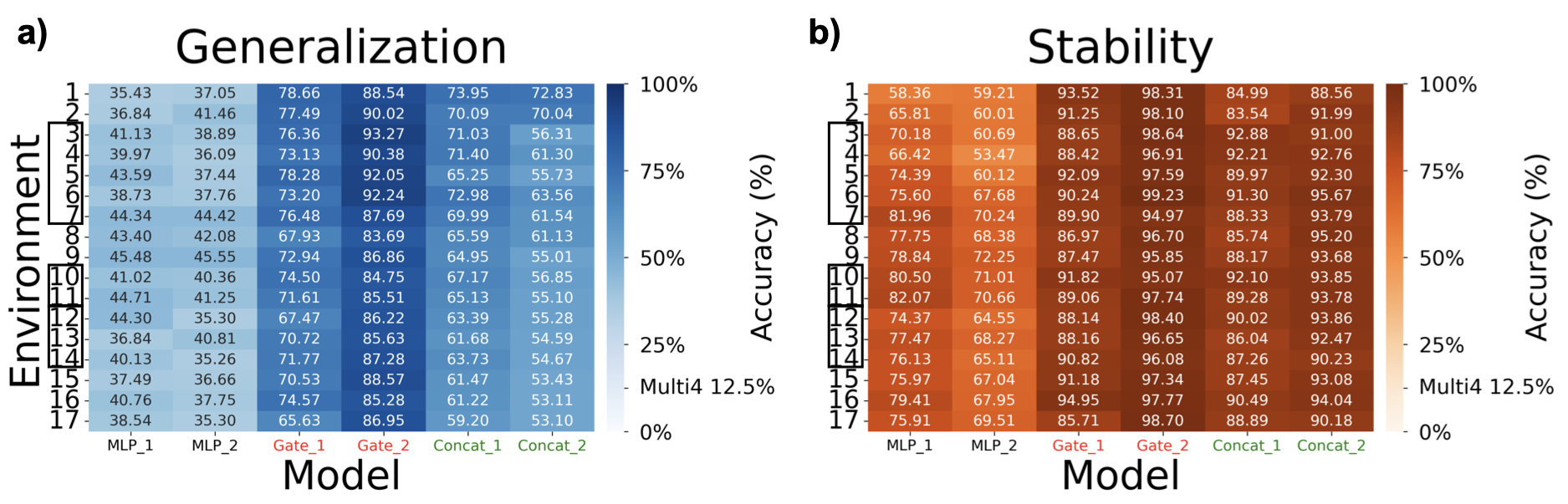}
\paragraph*{Fig. S3}
\label{S3_Fig}
\textbf{Accuracy across the 17 unique connected first-regime variants in the middle environment for all models.} (a) generalization accuracy; (b) stability accuracy. Attention-based models outperform MLPs on generalization and stability under all connected first regimes.  Regimes are ordered by increasing ASPL (higher to lower connectivity). Black boxes on the y-axis indicate identical ASPL (i.e., matched connectivity).

\end{document}